\documentclass[10pt,journal,compsoc]{IEEEtran}

\ifCLASSOPTIONcompsoc
  \usepackage[nocompress]{cite}
\else
  \usepackage{cite}
\fi
\usepackage{amsmath}
\usepackage{amssymb}
\usepackage{array}
\usepackage{booktabs}
\usepackage{capt-of}
\usepackage{graphicx}
\usepackage{makecell}
\usepackage{multirow}
\usepackage[T1]{fontenc}
\usepackage{algorithm}
\usepackage{algpseudocode}
\usepackage{url}
\usepackage[table]{xcolor}
\usepackage[hidelinks]{hyperref}

\graphicspath{{figures/}}

\begin{document}
\def\floatpagepagefraction{0.7}
\def\textpagefraction{.01}

\title{From Explanations to Interventions: Execution-Guided Counterfactual Synthesis in Temporal Graphs}
\author{Minwoo~Yu and Young-guk~Ha%
\IEEEcompsocitemizethanks{%
\IEEEcompsocthanksitem M. Yu and Y.-g. Ha are with the Smart Computing Laboratory, Department of Computer Science \& Engineering, Konkuk University, Seoul 05029, Republic of Korea.
E-mail: \{snowypainter, ygha\}@konkuk.ac.kr.
\IEEEcompsocthanksitem Y.-g. Ha is the corresponding author.}}

\markboth{From Explanations to Interventions}%
{Yu and Ha: Execution-Guided Counterfactual Synthesis in Temporal Graphs}

\IEEEtitleabstractindextext{%
\begin{abstract}
Can a trace explaining model execution also compute the changes needed for a specified alternative prediction? We propose trace-guided intervention search, which uses executable reasoning traces as an intermediate representation for intervention synthesis. A Specified-Foil Counterfactual edits past events so that a frozen temporal predictor selects a designated foil. Our method constructs facts and replacement values from completed original and foil executions and recovered unmet conditions. Proposal generation constructs edits and selects candidates within a fixed cap; exact replay verifies foil top-1 outcomes among retained edits and compositions. Implemented in LiFTER for continuous-time dynamic graphs (CTDGs) and TLogic for temporal knowledge graphs (TKGs), the method improves success over coordinate-based proposal generation by 13.7--34.7 percentage points on four CTDG datasets and 60.0--83.3 points on two TKG datasets under matched downstream search and a proposal cap of 32. Separate shared-candidate comparisons retain 85.7--93.6\% of black-box greedy's CTDG success rate with 75.0--80.0\% fewer predictor evaluations. A Pulse case study confirms simulator-level survival for five of six interventions. Executable traces thus provide both explanatory evidence and a reusable computational representation for constructing and testing specified alternatives.
\end{abstract}

}

\maketitle
\IEEEdisplaynontitleabstractindextext
\IEEEpeerreviewmaketitle

\section{Introduction}
\label{sec:introduction}

Explainable AI asks what evidence makes a model's prediction understandable~\cite{miller2019explanation}. If the reasoning used to produce a prediction is accessible, could that reasoning also compute the changes needed for a different outcome? We use \textbf{executable traces both as evidence for predictions and as an intermediate representation for intervention synthesis}. Our setting is temporal graphs, in which events and relations accumulate over time. Temporal knowledge graphs (TKGs) represent timestamped relational facts, whereas continuous-time dynamic graphs (CTDGs) represent continuous interaction streams. In both settings, we start from a frozen predictor's original prediction \(A\) and compute what must change in the past for a designated alternative \(B\) to be selected.

We formulate this question as a \textbf{Specified-Foil Counterfactual}. Specifying a foil is already established in contrastive explanation and targeted counterfactual research~\cite{wachter2018counterfactual,karimi2022recourse}. In temporal graphs, the computational challenge is to connect that objective to concrete past-event edits. Allowing changes to endpoints, relations, and timestamps alongside event deletion and insertion rapidly expands the space of edits and combinations. CoDy and TemGX search for event removals that change the original prediction using MCTS and structural--temporal selection, respectively~\cite{qu2025cody,lu2026temgx}. These methods reveal the model's dependence on evidence for the original. Our focus is \textbf{how to obtain concrete edit values for a designated outcome from predictor execution}, rather than assigning a new name to an existing objective. Search using output feedback alone assesses an edit's effect only after constructing and replaying it. The central computational question is which events and values to construct as promising candidates before replay.

Executable neuro-symbolic predictors expose structure for this computation. The original's completed execution identifies facts on which its support depends. The foil may also have completed support or suppression, and grounding a rule with the foil as its conclusion can recover unmet premises. A foil's failure to be selected therefore need not imply an absence of execution information. In a MOOC example, one interaction participated in transitions that both supported the original and suppressed the foil. Deleting it made the designated rank-5 content item top-1. In a TKG example, shifting the time of an intent-to-negotiate fact changed the predicted consultation partner (Section~\ref{subsec:qualitative}). Fact dependencies exposed by explanation thus identify where to intervene and which values to test.

We instantiate this connection through \textbf{diagnosis, synthesis, and verification} in trace-guided intervention search. Diagnosis identifies completed executions for the two candidates and recoverable unmet foil-side conditions. Synthesis converts these conditions into fact deletion or insertion and endpoint, relation, or time replacements, limiting candidates through model-specific priorities. Verification applies atomic edits and promising two-edit combinations to the history and checks whether the same frozen predictor selects \(B\) as top-1. LiFTER's signed interaction executions and TLogic's temporal rule groundings support this procedure while retaining their distinct score aggregations. We reuse the backbones' existing tracing and replay capabilities and place \textbf{a computation from execution conditions to concrete edits} between them.

Across four CTDG and two TKG datasets, we first compare proposal generation under identical downstream search (Section~\ref{sec:candidate-synthesis}). At a proposal cap of 32, success increases over coordinate-based generation by 13.7--34.7 percentage points in CTDGs and 60.0--83.3 points in TKGs. This evaluates proposal generation as the construction of edit candidates and selection of candidates to verify within the cap. Separate shared-candidate comparisons retain 85.7--93.6\% of black-box greedy's CTDG success rate with 75.0--80.0\% fewer predictor evaluations. End-to-end measurements across all six datasets yield speedups of 1.34--3.20. Bounded exhaustive comparisons and ablations examine missed successes and component roles. In a Pulse case study, all six returned treatment interventions are executed in separately initialized simulations, confirming the specified survival outcome in five cases (Section~\ref{sec:simulator-validation}).

Building on the targeted counterfactual, abduction, and executable temporal reasoning literature discussed in Section~\ref{sec:related_work}, we summarize our contributions as follows.
\begin{itemize}
\item \textbf{Execution-grounded proposal generation.} We transform executed evidence and recovered unmet conditions for the original and foil into concrete past-event edit candidates. The central connection computes which facts to modify and which values to substitute from grounded conditions.
\item \textbf{An explicit synthesis--verification procedure.} We combine a specified-foil objective, admissible edits, unit-cost bounded composition, and predictor-level verification, and specify adapters for LiFTER's and TLogic's distinct execution and aggregation structures.
\item \textbf{Evaluation in two temporal graph formalisms.} We analyze success and computational cost, execution information and priorities, operators and composition, stability, and constraint sensitivity for fixed proposals. A simulator case study demonstrates an application pathway for confirming or challenging returned interventions in a separate execution environment.
\end{itemize}

Beyond making explanations readable, the broader question is \textbf{what further computation the execution structure within an explanation enables}. Faithful reasoning linked to concrete facts can support understanding of a prediction while providing a reusable computational representation for constructing and testing alternatives.

\section{Related Work}
\label{sec:related_work}

\subsection{Counterfactual and Contrastive Explanation}

Contrastive explanation compares a fact \(A\) with a foil \(B\) to explain why \(A\) rather than \(B\), whereas counterfactual explanation seeks small input changes that shift a predictor's output to an alternative outcome~\cite{wachter2018counterfactual,miller2019explanation}. Algorithmic recourse extends this objective to computing actionable changes for desired outcomes, distinguishing plausible explanations from executable interventions~\cite{karimi2022recourse}. Both traditions can use foils, and desired-class formulations already exist in counterfactual research. We apply this targeted objective to temporal graphs and focus on converting executed groundings and recoverable unmet conditions into concrete edit values. The output is a temporal intervention that causes the frozen predictor to select \(B\) as top-1, rather than only a description or attribution of the difference between \(A\) and \(B\). Real-world causal counterfactuals additionally require a causal model and identification assumptions~\cite{pearl2009causality}.

\subsection{Counterfactual Reasoning in Knowledge Graphs}

Graph counterfactual explanation broadly studies prediction changes through additions or deletions of edges and features; surveys identify validity, proximity, sparsity, and plausibility as central evaluation dimensions~\cite{prado2024graphcf,guo2025counterfactual}. CF-GNNExplainer, for example, learns small subgraph perturbations through edge deletion to change node-classification predictions~\cite{lucic2022cfgnn}. Knowledge graph research has explored explicit fact interventions in several forms. Pezeshkpour et al. search for single-fact additions and deletions that change link predictions~\cite{pezeshkpour2019adversarial}; Imagine generates plausible triples to change a target triple's rank~\cite{barile2024imagine}. CFKGR adds hypothetical premises and determines which facts change under logical rules~\cite{zellinger2024cfkgr}. Abductive reasoning constructs logical hypotheses backward from observed conclusions~\cite{bai2024abduction}. These studies establish foundations for fact addition, target-directed modification, and inverse reasoning. In particular, constructing hypotheses backward from a conclusion is an established principle of abduction. Our contribution is not inverse reasoning itself, but a procedure that obtains editable facts and replacement values from a frozen temporal predictor's execution and connects them to bounded composition and predictor replay.

\subsection{Executable Reasoning in Temporal Knowledge Graphs}

TKGs represent facts as quadruples \((s,r,o,t)\), making relations, entity bindings, and time explicit objects of reasoning. TLogic extracts rules from temporal random walks and predicts future facts through time-consistent grounding~\cite{liu2022tlogic}; TILP learns recurrence, temporal order, intervals, and durations through differentiable temporal rules~\cite{xiong2023tilp}. TFLEX combines logical entity operators with temporal operators~\cite{lin2023tflex}, and TEILP predicts query-event times from rule-satisfying events and time intervals~\cite{xiong2024teilp}. INFER incorporates temporal validity and frequency into neural-symbolic rule application~\cite{li2025infer}. These approaches express temporal reasoning through symbolic or differentiable operators. We specifically reuse executable groundings from which concrete fact dependencies can be recovered, connecting completed and supported partial groundings to fact edits.

\subsection{Explanations for Continuous-Time Dynamic Graphs}

CTDG explainability has developed around identifying past events relevant to future-link predictions. T-GNNExplainer finds factual event subsets that preserve the original prediction~\cite{xia2023tgnnexplainer}, while TempME explains predictions through temporal motifs~\cite{chen2023tempme}. Among counterfactual methods, CoDy selects candidate events using spatio-temporal vicinity and heuristic policies, then uses MCTS to search for removal sets that change the original prediction~\cite{qu2025cody}. TemGX selects temporal subgraphs for removal using structural influence and time decay and verifies prediction changes~\cite{lu2026temgx}. CTM-Explainer conceptualizes event removal and synthetic addition, while its quantitative evaluation focuses on deletion-based event selection and changes in the original score~\cite{zhao2026ctm}.

T-GNNExplainer and TempME seek factual evidence preserving the original prediction; CoDy and TemGX seek counterfactual event removals that change it. CTM-Explainer's experiments likewise focus on changes in the original score. The latter objective corresponds to prediction invalidation without a prespecified replacement. A Specified-Foil Counterfactual fixes the alternative future link as input and requires it to become top-1. Respecting these distinct objectives, we study proposal construction and verification for a specified target. Our output-feedback controls on shared candidates compare selection costs under the same success criterion; they are not targeted reimplementations of the original CoDy or TemGX methods.

LiFTER provides grounded reasoning, signed contributions, and fact-level intervention and replay~\cite{yu2026lifter}. Building on these capabilities, we construct an adapter and bounded solver that synthesize editable facts and replacement values from original and foil conditions. The distinction is reuse of execution information for proposal computation between explanation and replay, rather than introducing explanation or replay for the first time.

\section{Problem and Execution Interface}
\label{sec:problem_definition}

We construct changes to the past that cause a frozen temporal predictor to select a specified alternative. Whereas prediction invalidation treats replacement of the original prediction as success, a Specified-Foil Counterfactual fixes that replacement as part of the problem input. Contrastive explanation broadly explains the grounds distinguishing two outcomes; here, we connect those grounds to executable edits.

\subsection{Temporal Prediction and Specified Foils}
\label{executable-temporal-prediction}

Let \(F_{<T_q}\) denote the history preceding query time \(T_q\), and let \(\mathcal Y_q\) be the complete candidate catalog. A candidate is a future destination for a CTDG query and an object for a TKG query \((s,r,?,T_q)\). The original prediction is determined by the frozen predictor's scores and a fixed tie-breaking rule.
\begin{equation}
\hat Y=\arg\max_{Y\in\mathcal Y_q}s(q,Y;F_{<T_q}).
\label{eq:original-prediction}
\end{equation}
The foil \(Y^{\mathrm{foil}}\) is a catalog element distinct from the original, specified before intervention search. Let \(\hat Y^\Delta\) denote the prediction after intervention. The two success criteria are
\begin{align}
\text{Prediction invalidation:}\quad&\hat Y^\Delta\ne\hat Y,\\
\text{Specified-Foil:}\quad&\hat Y^\Delta=Y^{\mathrm{foil}}.
\label{eq:success-conditions}
\end{align}
The criteria can coincide for binary prediction. In temporal link prediction with multiple destinations, the specified-foil criterion additionally requires selecting the designated candidate. We also write \(A=\hat Y\) and \(B=Y^{\mathrm{foil}}\).

\subsection{Foil Specification and Benchmark Construction}
\label{foil-specification-and-benchmark-construction}

In applications, the foil is determined by the downstream user or the analytical objective. Existing temporal benchmarks do not annotate user intent, so evaluation requires a foil-construction protocol. We first compare the ground-truth destination and historical best/recent alternatives, then adopt rank-based construction as the main benchmark to retain all evaluation queries while controlling the foil's model-relative position.
\begin{equation}
Y_k^{\mathrm{foil}}=\operatorname{Rank}_k
\{s(q,Y;F_{<T_q}):Y\in\mathcal Y_q\},\quad k\in\{2,5,10\}.
\label{eq:rank-foil}
\end{equation}
Rank is a controlled evaluation coordinate, not a definition of user utility or minimum edit distance. The construction comparison measures availability and original rank; it does not substitute for evaluating solver performance on every type of user-specified foil. The protocol is detailed in Section~\ref{queries-and-foils} and Supplementary Section~\ref{supp:foil-construction}.

\subsection{Admissible Edits and Optimization}
\label{contrastive-counterfactual-reasoning}
\label{intervention-space}

For CTDG interactions, DELETE/INSERT modify event existence, REWIRE modifies the endpoint, and SHIFT modifies the timestamp. TKG quadruples additionally contain a relation coordinate, which is modified by RELABEL.
\begin{align}
\mathcal O_{\mathrm{CTDG}}&=\{\mathrm{DELETE},\mathrm{INSERT},\mathrm{REWIRE},\mathrm{SHIFT}\},\\
\mathcal O_{\mathrm{TKG}}&=\mathcal O_{\mathrm{CTDG}}\cup\{\mathrm{RELABEL}\}.
\end{align}
REWIRE, RELABEL, and SHIFT replace one coordinate while preserving correspondence with the existing event. The same change can be expressed as DELETE followed by INSERT, but then incurs two atomic edits. Their roles therefore reflect the unit of cost as well as representational expressiveness.

A validity predicate \(V\) defines admissible changes to the past. Edits to existing facts are restricted to the pre-query history, and new timestamps must also precede the query. Entities and relations must belong to the predictor's domain, and conflicting edits to the same fact are prohibited. The target fact at the query time therefore cannot enter the history. This temporal restriction differs from prohibiting the target's endpoints and relation throughout the past. CTDG repair excludes direct source--foil insertion, whereas TKG prefix completion constructs past triples required by learned rules and does not categorically exclude a previous occurrence of the query's subject--relation--object triple. Value-construction rules are given in Section~\ref{subsec:synthesis}; additional history-support filtering is examined in Section~\ref{admissibility-sensitivity}.

Let \(\Delta\) be a set of interventions, \(F_{<T_q}^\Delta\) the edited history, and \(C(\Delta)\) its cost. The global objective is
\begin{align}
\Delta^*&\in\arg\min_\Delta C(\Delta)\\
\text{s.t.}\quad V(\Delta,F_{<T_q})&=1,\\
\arg\max_{Y\in\mathcal Y_q}s(q,Y;F_{<T_q}^{\Delta})&=Y^{\mathrm{foil}}.
\label{eq:specified-foil-objective}
\end{align}
Our solver and experiments use unit cost per atomic edit:
\begin{equation}
C(\Delta)=|\Delta|,\qquad |\Delta|\le2.
\label{eq:unit-cost}
\end{equation}
We distinguish the global objective from the algorithm's guarantees. The solver returns a successful intervention with the fewest edits among its finite generated candidates and bounded compositions; it does not establish global minimality over unexamined values or combinations. Failure means that no solution was found within the search budget. The bounded exhaustive comparison likewise evaluates minimality only within a fixed candidate set and budget.

Exact replay applies the same predictor to the edited history and checks its returned top-1 candidate. CTDG verification evaluates the complete catalog. The TLogic experiments use a fixed inference procedure with candidate-count stopping, which differs from ranking after applying all rules (Section~\ref{subsec:priority}). Predicate \(V\) determines syntactic and protocol validity, whereas replay determines model-level success. The real-world plausibility of a syntactically admissible INSERT or REWIRE requires additional domain constraints or expert judgment.

\subsection{Execution Interface}
\label{subsec:execution-interface}

The problem and replay criterion also apply to black-box predictors. Our synthesis procedure additionally requires the following execution information.
\begin{enumerate}
\item \textbf{Faithful execution.} The trace identifies rules or grounded computations that actually contribute to candidate scores.
\item \textbf{Addressable grounding.} The facts, entity bindings, and temporal conditions used in execution can be identified and linked to concrete edits.
\item \textbf{Foil-queryable execution.} Completed computations for a designated candidate can be inspected, and partial groundings and remaining conditions can be recovered for supported rules.
\item \textbf{Intervention-closed replay.} The same inference procedure can be applied to an edited history to recompute groundings and candidate scores.
\end{enumerate}
This information may be exposed through a dedicated API or recovered by an adapter from existing rules and the inference engine. It is unnecessary to store every failed proof or enumerate all missing conditions. The supported conditions determine the range of candidates that synthesis can construct.

We implement adapters for LiFTER's signed interaction executions and TLogic's time-consistent groundings. Architectures that explicitly represent temporal rules and variable groundings, such as TILP, may also provide relevant information~\cite{xiong2023tilp}. This does not imply that adapter feasibility or performance has been verified on those backbones; our quantitative evaluation covers LiFTER and TLogic.

\section{Execution-Guided Intervention Synthesis}
\label{sec:method}

\begin{figure*}[t]
\centering
\includegraphics[width=\textwidth]{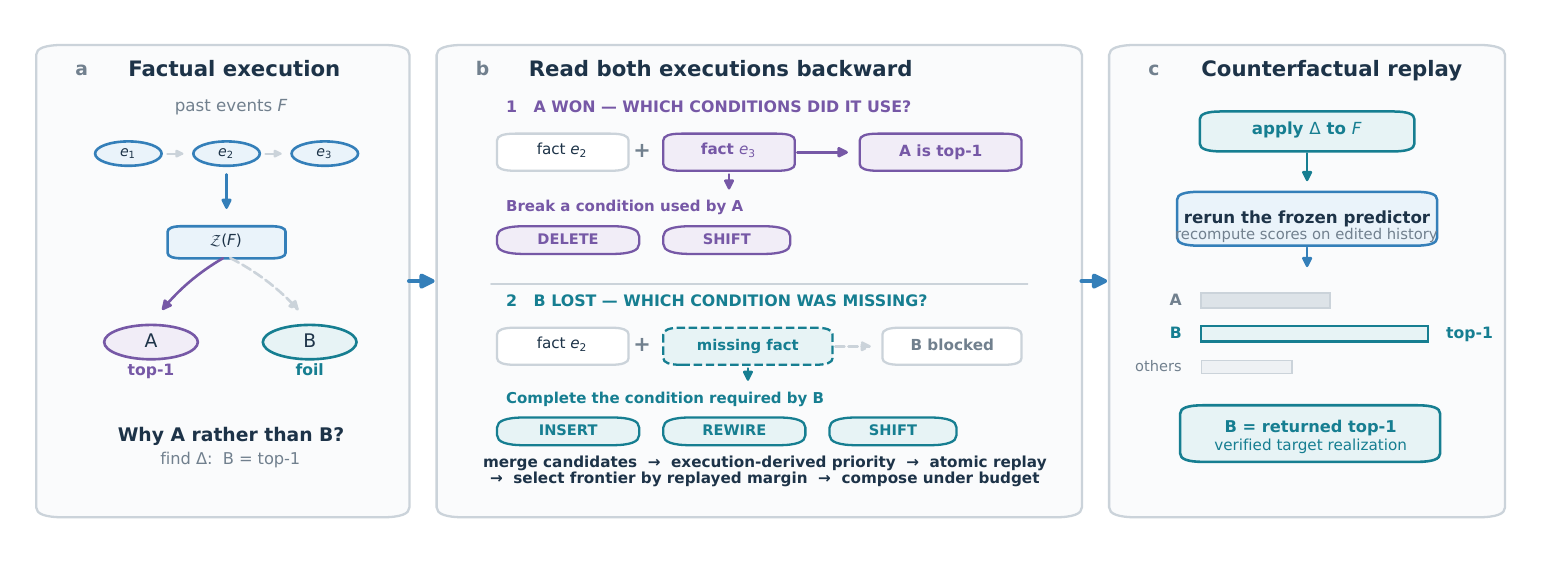}
\caption{From execution to intervention. Candidates are constructed by modifying original support or completing foil-side premises, and verified by exact replay. The figure illustrates a representative partial-grounding path; for signed predictors, releasing already-executed foil suppression also contributes to proposal generation.}
\label{fig:overview}
\end{figure*}

Our method uses predictor execution information as an intermediate representation for intervention synthesis. Diagnosis identifies completed executions and partial groundings that motivate edits; synthesis maps their conditions to concrete event edits; verification checks the resulting prediction changes. Fig.~\ref{fig:overview} illustrates representative paths through this procedure. A foil may have completed support and suppression as well as unmet conditions.

\subsection{Diagnosis: Completed and Partial Executions}
\label{subsec:diagnosis}

Let \(\mathcal T_A,\mathcal T_B\) denote the completed traces of the original \(A\) and foil \(B\), and let \(\widetilde{\mathcal T}_B\) denote partial groundings recovered with the foil fixed as the conclusion. Completed traces preserve the facts and bindings that actually contributed to score computation. A partial grounding identifies the remaining conditions after some rule premises have been satisfied. We do not assume that the predictor automatically stores every failed execution; the adapter recovers partial groundings from existing rules and history.

We retain each model's aggregation rather than replacing it with a common signed sum. Executed rule or computation values are combined by the model-specific aggregation function, and diagnosis uses the score difference between the two predictions:
\begin{align}
s(q,Y;F)&=\operatorname{Agg}_{f}\bigl(Y,\mathcal T_Y(F)\bigr),\\
D(F)&=s(q,A;F)-s(q,B;F).
\label{eq:execution-gap}
\end{align}
Hereafter, \(F\) denotes the pre-query history. LiFTER's signed contributions identify facts that support \(A\) or suppress \(B\). TLogic aggregates nonnegative rule scores through noisy-OR, so the same negative-contribution interpretation does not apply. Its adapter instead uses active original groundings and rule-body prefixes leading toward the foil.

LiFTER's binary-rule repair considers length-two rules with positive effective weights for which exactly one of the two body atoms is valid. It binds the query source and foil to \(X,Y\), respectively, obtains the intermediate entity \(Z\) from the satisfied atom, and determines the other atom's endpoints. Cases with unresolved roles are excluded. The TLogic adapter executes the prefix preceding the last atom of a rule for the query relation. The prefix's terminal entity, last timestamp, and required relation determine a candidate final fact ending at the foil. A length-one rule uses an empty prefix. The prefix does not prove that the corresponding triple is absent throughout the history; it provides a starting point for constructing a terminal condition leading to the foil. These are adapter-supported completion procedures, not exhaustive recovery of arbitrary multiple-premise failures.

\subsection{Synthesis: From Conditions to Concrete Edits}
\label{subsec:synthesis}

The conditions identified during execution specify which past events to delete or modify, or which events to add. For example, if a TLogic rule has reached entity \(z\) and requires a final relation \(r_\ell\) leading to foil \(B\), synthesis constructs the following candidate fact:
\begin{equation}
(z,r_\ell,B,t'),\qquad t'<T_q.
\label{eq:grounding-to-fact}
\end{equation}
The adapter can INSERT this fact, REWIRE an existing fact with the same relation toward \(B\), or RELABEL a fact with the same endpoints to \(r_\ell\). It constructs a SHIFT when the timing of an existing binding is unsuitable. Replay determines whether the resulting rule satisfies its temporal and variable constraints. Table~\ref{tab:condition-edit} summarizes the mappings implemented by the two adapters.

\begin{table*}[t]
\centering
\caption{Grounded conditions and the concrete edits constructed by each adapter. Values are proposals; exact replay determines their actual effect.}
\label{tab:condition-edit}
\small
\begin{tabular}{p{0.09\textwidth}p{0.265\textwidth}p{0.275\textwidth}p{0.27\textwidth}}
\toprule
Operator & Execution condition & LiFTER / CTDG construction & TLogic / TKG construction\\
\midrule
DELETE & Original support, or signed foil suppression, uses an addressed fact & Remove the fact; accumulate signed pressure & Remove a fact in an active original walk\\
INSERT & A body atom is missing; its variables can be bound & Instantiate the missing endpoints at recent grid times; also use inverse-transition endpoints & Append the required last relation from a valid prefix toward $B$ at $T_q-1$\\
REWIRE & A present fact can supply the required binding & Replace an outgoing fact's destination by the missing/transition endpoint & Redirect the terminal original fact or a prefix-outgoing fact toward $B$\\
RELABEL & Endpoints match but the required relation differs & Not applicable to the interaction representation & Replace that relation by the last body relation\\
SHIFT & Temporal order or recency can change an execution & Keep endpoints/features; replace time with a source-history grid value & Cross an original walk's time boundary, or move a matching foil fact to $T_q-1$\\
\bottomrule
\end{tabular}
\end{table*}

CTDG timestamp candidates form a finite grid comprising midpoints between recent distinct event times for the query source and a representable time immediately before the query. Near-miss INSERT uses up to the last two grid times, whereas transition INSERT uses the most recent time. REWIRE preserves the existing event's timestamp and features; SHIFT preserves its endpoints and features. INSERT uses a feature prototype from an influential historical fact, or a zero vector when no such fact is available. These rules restrict the unbounded continuous timestamp and feature spaces to searchable candidates.

LiFTER's ordered transitions provide another synthesis path. With the immediately preceding destination fixed, the adapter computes the difference in first-/second-order transition contributions to \(B-A\) for each candidate historical endpoint. For top endpoints with positive differences, it constructs INSERT, REWIRE of a recent interaction, and SHIFT of a matching interaction. This inverts transition contributions; it does not compute the exact post-edit score of the entire predictor in advance.

TLogic uses \(T_q-1\) as the completion time. SHIFT candidates from original groundings cross a neighboring fact's temporal boundary by one unit or move the last fact to an earlier time. Both adapters thus construct prescribed time candidates and determine their effects through replay. TKG edits also update the corresponding official inverse facts.

\subsection{Pre-replay Priority and Model-specific Aggregation}
\label{subsec:priority}

Whether an edit actually reduces the original--foil gap is determined by executing the edited history. We distinguish this post-replay effect from the priority used to select candidates beforehand:
\begin{equation}
R(\delta)=D(F)-D(F^\delta).
\label{eq:expected-gain}
\end{equation}
Before replay, we use a separate execution-derived priority \(\widehat R\). It is neither the exact value of \(R\) nor a guaranteed bound, but a heuristic based on the following model-specific information.

In LiFTER, a fact's pressure sums its positive original contributions and the absolute values of its negative foil contributions. DELETE receives this pressure, and SHIFT of the same fact receives half of it. Near-miss INSERT and REWIRE use the rule's positive effective weight; transition repair uses the positive \(B-A\) transition contribution computed above. When multiple paths generate the same edit, their priorities are summed and the edit is deduplicated.

TLogic uses the original trace score plus rule confidence, minus a body-position term \(0.01i\), as its base priority. Original SHIFT receives an offset of \(-0.05\), and terminal REWIRE receives \(+0.5\). Foil-prefix completion uses \(2+\text{confidence}\), with additional offsets of \(0.2,0.1,0.15\) for REWIRE, RELABEL, and SHIFT, respectively. These coefficients are fixed heuristics chosen intuitively during implementation, without adjustment based on test results. They are neither learned during inference nor adapted per query. Duplicate edits retain their highest priority.

Priority is also distinct from the predictor score. LiFTER uses signed grounded contributions in an additive score. TLogic evaluates each rule's matching walks for a candidate using the official confidence--recency scoring function and aggregates the resulting rule-level values through noisy-OR~\cite{liu2022tlogic}:
\begin{align}
s_{\mathrm{LiFTER}}(Y;F)&=b_Y+\sum_z\alpha_z(Y;F),\\
s_{\mathrm{TLogic}}(Y;F)&=1-\prod_r\bigl(1-u_{r,Y}(F)\bigr).
\label{eq:model-aggregations}
\end{align}
Individual walks in a TLogic trace are not treated as independent additive terms. The implementation reuses official rule application and scoring. In our experiments, candidate-count stopping halts further rule application once at least 10 candidates with distinct accumulated scores have been obtained. This does not truncate the candidate set to exactly 10, but neither does it guarantee that subsequent rules would leave scores or ranks unchanged. TLogic results therefore verify the \emph{top-1 returned by the fixed predictor}, including this stopping rule. They do not establish top-1 exactness after exhausting all rules or full-catalog verification. The same stopping configuration is used for the original prediction, foil selection, and edited replay.

\subsection{Verification: Atomic Replay and Bounded Composition}
\label{subsec:verification}

After validity filtering and deduplication, we retain up to \(K\) atomic edits with the highest priorities. Each is replayed, a frontier of \(H\) promising edits is formed, and nonconflicting two-edit pairs are evaluated. A successful atomic edit is preferred as a cost-one solution; otherwise, a successful pair is returned. Evaluation implementations for coverage and ablation also examine pairs after atomic success to record outcomes at both cost levels.

Reducing the original--foil gap differs from satisfying the final success criterion. The margin against the foil's strongest competitor in the complete catalog is
\begin{equation}
M_B(F^\Delta)=s(q,B;F^\Delta)
-\max_{Y\in\mathcal Y_q\setminus\{B\}}s(q,Y;F^\Delta).
\label{eq:target-margin}
\end{equation}
The CTDG frontier uses the replayed competing-candidate margin, whereas the TKG frontier uses replayed \(s_B-s_A\). Final success requires that the frozen predictor return \(B\) as top-1 under its tie-breaking rule. CTDG confirms this over the complete catalog; TKG uses the output under the stopping configuration specified above. Thus, \(s_B>s_A\) alone is insufficient for success. CTDG's Top-32-plus-foil screening is a batching device, and candidates passing screening are verified against the complete catalog.

Let \(\mathcal A_K\) be the atomic set and \(\mathcal P_H\) the nonconflicting frontier pairs. The examined space and return criterion are
\begin{align}
\mathcal S_{K,H}&=\mathcal A_K\cup\mathcal P_H,\\
\widehat\Delta&\in\arg\min_{\substack{\Delta\in\mathcal S_{K,H}\\
V(\Delta,F)=1,\;\hat Y^\Delta=B}}|\Delta|.
\label{eq:bounded-search}
\end{align}
Checking single edits first provides this minimum-edit property only under unit costs. Our experiments use budget 2 and do not perform general cost optimization with different operator costs. The number of distinct interventions that can be examined after proposal generation is bounded by
\begin{equation}
|\mathcal S_{K,H}|\le K+\binom{H}{2}.
\label{eq:search-bound}
\end{equation}
This is not a runtime bound accounting for trace recovery, proposal generation, screening, and individual replay costs. End-to-end cost is measured separately in Section~\ref{end-to-end-efficiency}. Failure indicates that no solution was found in this space, not global unreachability.

\subsection{Algorithm}
\label{algorithm-and-model-requirements}

Algorithm~\ref{alg:trace-search} includes the condition recovery and value construction performed by each adapter. A new backbone need not implement all operations internally: an adapter can recover the information in Section~\ref{subsec:execution-interface} from its existing inference procedure.

\begin{algorithm}[t]
\caption{Execution-guided intervention synthesis (unit cost, budget 2)}
\label{alg:trace-search}
\small
\begin{algorithmic}[1]
\Require frozen predictor $f$, history $F$, query $q$, foil $B$, caps $K,H$
\State scores, completed groundings $\gets f(F,q)$; $A\gets\arg\max$ scores
\State $\Gamma\gets\varnothing$
\For{facts supporting $A$ or suppressing $B$}
  \State add DELETE and grid/boundary SHIFT candidates to $\Gamma$
  \State in TKG, also redirect the terminal binding toward $B$
\EndFor
\For{supported partial groundings with conclusion $B$}
  \State bind source, foil and intermediate entities from the present atoms
  \State form missing fact endpoints/relation and admissible time candidates
  \State add INSERT and matching REWIRE, RELABEL or SHIFT to $\Gamma$
\EndFor
\State in LiFTER, add positive inverse-transition endpoint repairs
\State filter $\Gamma$ by $V$; deduplicate; retain top-$K$ by adapter priority
\State replay all retained atomic edits with the frozen predictor
\State form top-$H$ frontier by the adapter's replayed margin
\State replay nonconflicting frontier pairs; check the frozen predictor's returned top-1
\State \Return a successful minimum-edit candidate with its replayed trace,
\Statex \hspace{\algorithmicindent}or \textsc{NotFoundWithinSearchBudget}
\end{algorithmic}
\end{algorithm}

LiFTER's existing traces, signed contributions, fact interventions, and replay~\cite{yu2026lifter} provide the computational foundation. Our contribution is to generate concrete candidate values from completed/partial conditions and connect bounded synthesis to verification for a specified foil. The TLogic adapter implements the same connection while preserving its existing forward grounding and aggregation.

\section{Experiments}
\label{sec:experiments}

We evaluate the computational value and scope of reusing execution information to compute edits. First, we compare proposal-generation procedures under identical downstream search in CTDGs and TKGs. We then compare success, evaluation counts, and end-to-end time on shared candidates. Bounded exhaustive comparisons and ablations examine missed successes and component roles, while diagnostics assess candidate spaces, foils, and admissibility. Qualitative cases show which executions the synthesized edits change. Simulator application is examined separately in Section~\ref{sec:simulator-validation}.

\subsection{Experimental Setup}\label{experimental-setup}

CTDG experiments use LiFTER and the public Wikipedia, Reddit, MOOC, and LastFM interaction datasets~\cite{kumar2019jodie}. For each dataset, we use the most recent 70,000 events, train LiFTER on the first 85\%, and evaluate on the remaining 15\%. Each query receives only preceding events; its candidate catalog contains all destinations appearing in the same 70,000-event window. TKG experiments use ICEWS14 and ICEWS18 timestamped relational facts~\cite{boschee2015icews} and the official TLogic implementation~\cite{liu2022tlogic}. CTDG foil selection and replay verification use the complete destination catalog rather than sampled negatives. TKG foils are selected from rankings returned by TLogic with candidate-count stopping set to 10, and edited histories use the same inference configuration. This differs from verifying a full-catalog ranking obtained by exhausting all rules.

\paragraph{Frozen predictor}\label{frozen-predictor}

LiFTER is trained independently of counterfactual search and frozen for all interventions. Main results use seed 7; stability is evaluated with independently trained checkpoints using seeds 7, 17, and 37. Hyperparameters, backbone settings, and bootstrap procedures are provided in Supplementary Section~\ref{supp:implementation} and Tables~\ref{tab:backbone-settings}--\ref{tab:stability}.

\paragraph{Queries and foils}\label{queries-and-foils}

The main diagnostic and trace-efficiency experiments select 1,000 evenly spaced rows across each dataset's 10,500-row evaluation stream. For each row, the source and query timestamp are fixed. The frozen predictor's pre-intervention rank-1 candidate is the original, and its rank-2, rank-5, and rank-10 candidates are the three foils. This yields 3,000 original--foil comparisons per dataset and 12,000 in total. Foils are fixed before search and never replaced based on intervention outcomes.

The more computationally demanding bounded exhaustive comparisons and ablations use 100 queries per dataset, fixed before inspecting results. All ablations share the same query subset, and empty candidate sets count as failures.

For each TKG query \((s,r,?,t)\), the pre-intervention rank-1 object is the original and the rank-2/5/10 objects are the foils. Each of ICEWS14 and ICEWS18 contributes 300 comparisons from 100 queries and three foils.

Before adopting rank-controlled foils, we compared ground-truth destinations and two historical constructions on 100 queries from each of the four CTDG datasets. Definitions, availability, and median-rank results are given in Supplementary Section~\ref{supp:foil-construction} and Table~\ref{tab:foil-construction}.

\begin{figure}[t]
\centering
\includegraphics[width=0.96\columnwidth]{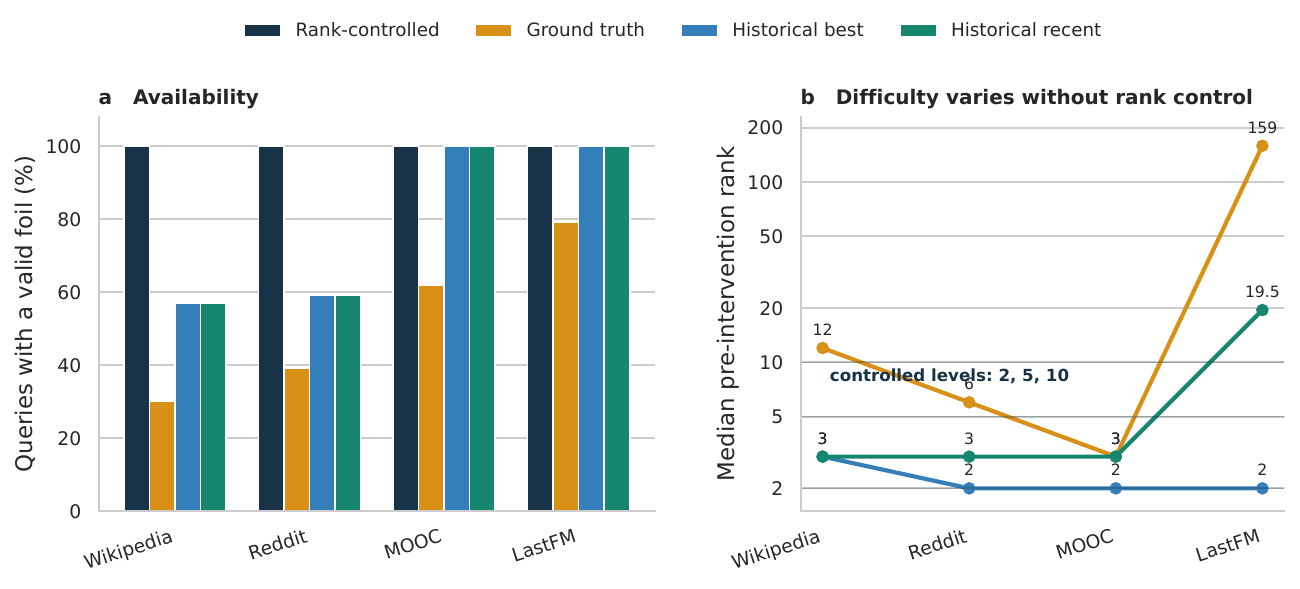}
\caption{Ground-truth and historical foils are defined for only some queries, with substantially different difficulty across datasets. Rank-controlled foils retain all queries while fixing comparison positions at ranks 2, 5, and 10.}
\label{fig:foil-construction}
\end{figure}

Fig.~\ref{fig:foil-construction} and Supplementary Table~\ref{tab:foil-construction} show that ground-truth and historical constructions are available for only some queries and vary in difficulty across datasets. Rank-2/5/10 retains all queries while fixing positions within model-induced difficulty. Application foils are not restricted by rank; the solver accepts any destination in the catalog.

\subsection{Proposal Generation under Matched Search}
\label{sec:candidate-synthesis}

We evaluate whether execution-derived candidates themselves improve access to the specified foil. Each of the four CTDG and two TKG datasets uses 100 fixed queries with rank-2/5/10 foils. CTDG uses seed-7 LiFTER, and TKG uses the same frozen TLogic as the other experiments. Coordinate-based generation constructs endpoint and timestamp alternatives, and TKG relation alternatives, from the query, historical events, and specified foil without inspecting learned rules or execution traces. Execution-grounded generation constructs editable events and values from completed and incomplete executions and retains candidates according to the existing execution priority. Both procedures use the same edit operators, admissibility conditions, budget of at most two edits, and atomic proposal cap $K\in\{4,8,16,32\}$. Retaining the first $K$ candidates in each procedure's ordering makes the proposal sets nested across caps.

Downstream search is identical once candidates have been selected within the cap. All atomic candidates are replayed; up to eight with the largest foil-minus-original score differences form the frontier, and nonconflicting two-edit combinations are evaluated. Success requires that the designated foil be top-1 over the complete CTDG destination catalog or the top-1 returned by the TKG predictor with candidate-count stopping 10. Identical candidates and combinations share replay outcomes, and empty sets count as failures. Here, proposal generation comprises constructing edits and selecting candidates for verification within the cap before replay. Only this proposal-generation procedure varies; the downstream selector and verification criterion remain fixed.

\begin{table*}[t]
\centering
\caption{CTDG and TKG target success (\%) under matched downstream search. Each cell reports coordinate-based / execution-grounded proposal generation. $K$ is the maximum number of atomic proposals.}
\label{tab:candidate-synthesis}
\small
\setlength{\tabcolsep}{3pt}
\begin{tabular}{crrrrrr}
\toprule
$K$ & Wikipedia & Reddit & MOOC & LastFM & ICEWS14 & ICEWS18 \\
\midrule
4  & 4.3 / 4.7 & 5.0 / 11.3 & 16.3 / 34.0 & 10.3 / 7.7 & 4.7 / 38.7 & 2.3 / 48.3 \\
8  & 6.0 / 11.3 & 7.0 / 19.3 & 25.3 / 54.0 & 17.0 / 31.3 & 7.3 / 50.0 & 3.7 / 65.0 \\
16 & 8.3 / 19.7 & 7.3 / 26.3 & 30.3 / 63.7 & 23.7 / 41.7 & 8.0 / 65.7 & 5.3 / 86.7 \\
32 & 8.7 / 22.3 & 7.7 / 34.3 & 33.7 / 68.3 & 28.3 / 50.3 & 14.7 / 74.7 & 8.0 / 91.3 \\
\bottomrule
\end{tabular}
\end{table*}

In Table~\ref{tab:candidate-synthesis}, execution-grounded generation at $K=32$ improves success over coordinate-based generation by 13.7, 26.7, 34.7, and 22.0 percentage points on Wikipedia, Reddit, MOOC, and LastFM, respectively. Paired-bootstrap 95\% intervals, resampling the three foils of each query together, are [9.0, 18.7], [20.7, 33.0], [27.7, 41.3], and [15.0, 29.0] points (10,000 replicates; pointwise intervals). Execution-grounded generation has higher success on all six datasets at $K=8,16,32$. At $K=4$, however, LastFM success decreases from 10.3\% to 7.7\%, indicating that a very narrow cap can exclude useful edits during candidate selection.

Short proposal sets are not replenished from the other generator. At $K=32$, execution-grounded generation retains a mean of 31.4--32.0 candidates across CTDG datasets, 30.3 on ICEWS14, and 31.1 on ICEWS18. The results show that execution-grounded generation supplies candidates reaching more specified alternatives under the same cap and search. This experiment evaluates proposal generation under matched search; Section~\ref{trace-guided-intervention-search-efficiency} compares selection and screening costs with a shared candidate set.

TKG results likewise distinguish the generators under the same search. At $K=32$, success increases from 14.7\% to 74.7\% on ICEWS14 and from 8.0\% to 91.3\% on ICEWS18. Paired-bootstrap 95\% intervals for the increases are [51.3, 68.3] and [77.3, 88.7] percentage points, respectively. Execution-grounded generation has higher success at all four proposal caps on both TKG datasets.

\subsection{Search Efficiency on Shared Candidates}
\label{trace-guided-intervention-search-efficiency}

We compare trace-guided intervention search with random selection, event-recency-based locality selection, and black-box greedy, which prioritizes atomic candidates through replay. All controls share the original, foil, edit budget, and predictor-level success criterion; the candidate space also includes trace-generated candidates. This is therefore a \textbf{comparison of selection and screening costs under shared candidate access}. It does not compare proposal generation against a complete end-to-end solver that constructs candidates without traces.

We evaluate 3,000 comparisons per CTDG dataset from 1,000 queries and three foils, and 300 comparisons per TKG dataset. The main efficiency comparison uses at most 32 atomic candidates and composition beam 8 for CTDG, and 16 candidates and beam 4 for TKG. TKG search-width sensitivity is evaluated separately up to candidate cap 32 and beam 16 (Supplementary Table~\ref{tab:ablation-width}). Evaluation counts include all black-box greedy atomic screening. CTDG greedy uses the foil margin against competing candidates; TKG greedy uses the foil-minus-original score difference.

\begin{table*}[t]
\centering
\caption{Target success and predictor-evaluation reduction of trace-guided intervention search.}
\label{tab:search-efficiency}
\small
\setlength{\tabcolsep}{3.5pt}
\resizebox{\textwidth}{!}{%
\begin{tabular}{lrrrr}
\toprule
Formalism / dataset & Trace-guided intervention search & Black-box greedy & Success-count ratio & Evaluation reduction \\
\midrule
CTDG / Wikipedia & 30.8\% & 33.4\% & 92.3\% & 76.3\% \\
CTDG / Reddit & 37.1\% & 39.7\% & 93.5\% & 80.0\% \\
CTDG / MOOC & 77.2\% & 82.4\% & 93.6\% & 77.2\% \\
CTDG / LastFM & 57.0\% & 66.6\% & 85.7\% & 75.0\% \\
TKG / ICEWS14 & 64.0\% & 67.3\% & 95.0\% & 60.0\% \\
TKG / ICEWS18 & 85.7\% & 86.3\% & 99.2\% & 59.4\% \\
\bottomrule
\end{tabular}%
}
\end{table*}

\begin{figure}[t]
\centering
\includegraphics[width=0.96\columnwidth]{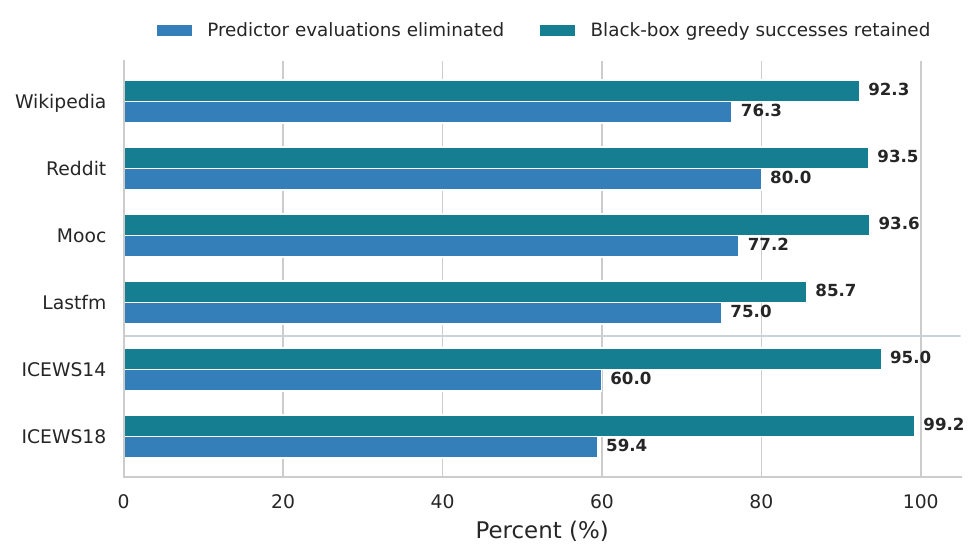}
\caption{Evaluation reduction and success-count ratio relative to greedy on shared candidates. The success-count ratio differs from recovery of the same successful cases (Eq.~\eqref{eq:success-ratios}).}
\label{fig:search-efficiency}
\end{figure}

In Table~\ref{tab:search-efficiency}, the proposed CTDG method attains 85.7--93.6\% of greedy's success rate with 75.0--80.0\% fewer evaluations. The corresponding TKG ranges are 95.0--99.2\% and 59.4--60.0\%. This success-rate ratio compares success counts, rather than recovery of the \emph{same cases} solved by greedy.
\begin{align}
\text{Success-count ratio}&=\frac{|S_{\mathrm{ours}}|}{|S_{\mathrm{greedy}}|},\\
\text{Paired success recovery}&=\frac{|S_{\mathrm{ours}}\cap S_{\mathrm{greedy}}|}
{|S_{\mathrm{greedy}}|}.
\label{eq:success-ratios}
\end{align}
The first metric is used in Table~\ref{tab:search-efficiency} and Fig.~\ref{fig:search-efficiency}; the second is used in the paired end-to-end evaluation below.

These results show that execution-derived priorities can replace a substantial portion of screening that first replays candidates to assess their effects. On ICEWS14/18, random selection achieves 54.3/69.7\% success, locality achieves 55.7/63.0\%, and the proposed method achieves 64.0/85.7\%. Under shared candidate access, trace-informed prioritization is more effective than recency or random selection. This comparison does not establish inferior performance of existing prediction-invalidation explainers or the impossibility of black-box targeted search.

\subsection{End-to-End Efficiency}\label{end-to-end-efficiency}

To determine whether fewer predictor evaluations translate into lower runtime, we measure end-to-end wall-clock time from proposal generation through exact replay. All methods use the same queries, candidate space, and hardware settings. Full measurement procedures and results are reported in Supplementary Section~\ref{supp:e2e-efficiency} and Table~\ref{tab:e2e-efficiency}.

\begin{figure*}[t]
\centering
\includegraphics[width=0.94\textwidth]{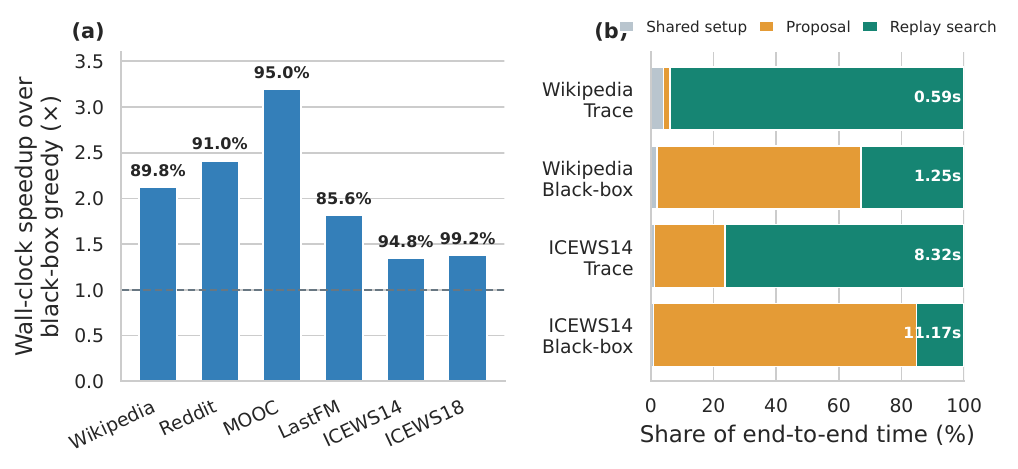}
\caption{End-to-end efficiency. (a) Mean wall-clock speedup over black-box greedy; labels above bars indicate the proportion of greedy-successful cases also solved by the proposed method. (b) Runtime breakdowns for representative CTDG and TKG datasets, distinguishing black-box screening over shared candidates from trace construction in the proposed method.}
\label{fig:e2e-efficiency}
\end{figure*}

In Fig.~\ref{fig:e2e-efficiency}(a), the proposed method is faster than black-box greedy on all six datasets, with speedups of 1.34--3.20$\times$. It recovers 85.6--99.2\% of the same cases solved by greedy, or 1,009 of 1,082 cases overall (93.3\%). Fig.~\ref{fig:e2e-efficiency}(b) locates the difference in the proposal stage. Mean runtime decreases from 1.254 to 0.590 seconds on Wikipedia and from 11.171 to 8.317 seconds on ICEWS14. The same trend holds for p95 latency on every dataset. Locality and random selection have lower proposal costs but lower target success. Execution traces thus offer an efficiency--effectiveness operating point that reduces black-box screening while retaining high target success.

\paragraph{Statistical stability}\label{statistical-stability}

Repeating CTDG experiments with three independently trained LiFTER checkpoints yields target-success standard deviations of 0.45--2.38 percentage points, with method orderings preserved across seeds. Paired cluster-bootstrap 95\% confidence intervals against random and locality exclude zero on all four datasets. Per-seed results and tests are reported in Supplementary Table~\ref{tab:stability}.

\subsection{Comparison with Exhaustive Search}\label{comparison-with-exhaustive-search}

To assess whether trace-guided intervention search achieves speed by missing many solutions, we exhaustively execute interventions in a small search space. For each prediction, rank-2, rank-5, and rank-10 candidates serve as separate foils. We test every atomic candidate and every nonconflicting two-candidate combination. This finite space contains at most 32 atomic candidates for CTDG and 16 for TKG. An edited temporal graph is successful only if the frozen predictor returns the specified foil as top-1; the successful intervention with the fewest edits is the minimum within this space.

Trace-guided intervention search receives the same foil, candidate space, and two-edit budget. Exhaustive search evaluates every admissible intervention, whereas trace-guided search uses original--foil trace differences to select at most 16 candidates and promising combinations. We measure the fraction of exhaustive-successful cases also solved by trace-guided search, recovery of the same minimum edit count, and reduction in evaluated interventions.

CTDG uses 300 comparisons per dataset from 100 queries and three foils. Each atomic edit and pair receives one shared replay verdict; results and frontier scores for the proposed method's selected subset are read from this common evaluation. The experiment therefore compares coverage and minimum-cost recovery within the same finite space, not wall-clock times of separately executed methods. Evaluation reduction is computed per comparison from the number of interventions selected for examination by each search and then averaged. It differs from the total physical cost of the shared audit.
\begin{table*}[t]
\centering
\caption{Comparison with exhaustive search in the fixed bounded candidate space.}
\label{tab:exhaustive}
\small
\setlength{\tabcolsep}{3.5pt}
\resizebox{\textwidth}{!}{%
\begin{tabular}{lrrrrr}
\toprule
Formalism / dataset & Exhaustive success & Trace-guided intervention search & Solution recovery & Minimum-cost recovery & Search reduction \\
\midrule
CTDG / Wikipedia & 25.3\% & 22.7\% & 89.5\% & 89.5\% & 91.5\% \\
CTDG / Reddit & 38.7\% & 36.0\% & 93.1\% & 93.1\% & 91.5\% \\
CTDG / MOOC & 79.3\% & 69.7\% & 87.8\% & 87.0\% & 91.6\% \\
CTDG / LastFM & 66.3\% & 55.3\% & 83.4\% & 79.4\% & 91.2\% \\
TKG / ICEWS14 & 71.2\% & 69.7\% & 97.9\% & 97.9\% & 83.1\% \\
TKG / ICEWS18 & 85.3\% & 85.3\% & 100.0\% & 100.0\% & 83.5\% \\
\bottomrule
\end{tabular}%
}
\end{table*}

Table~\ref{tab:exhaustive} reports the finite-space comparison. Recovery is measured over query--foil comparisons solved by exhaustive search, not over all possible interventions. Minimum-cost recovery likewise counts successes matching the exhaustive minimum edit count for the same comparison.

In CTDG, exhaustive search solves 76, 116, 238, and 199 cases on Wikipedia, Reddit, MOOC, and LastFM, respectively. The proposed method recovers 68, 108, 209, and 166, yielding recovery rates of 89.5, 93.1, 87.8, and 83.4\%. All proposed-method successes are subsets of exhaustive successes. Matching minimum edit counts are recovered in 68, 108, 207, and 158 cases; in two MOOC and eight LastFM cases, the method finds a two-edit solution instead of the exhaustive one-edit solution. Minimum-cost recovery also uses exhaustive successes as its denominator. The selected search space is 91.2--91.6\% smaller than the exhaustive space. This recovers a substantial portion of successes in a common bounded space through narrow selection, rather than providing a guarantee over all possible edits.

Minimality here is restricted to the fixed atomic candidates and at most two edits. The experiment measures how many fewer evaluations trace-guided search needs to recover answers obtained by examining every intervention in the same finite space.

\subsection{Ablation Study}\label{ablation-study}

Using 1,200 comparisons across four CTDG datasets and 300 per TKG dataset, we distinguish execution information used for proposal construction from post-replay selection objectives. All variants share queries and exact-replay verdicts. Table~\ref{tab:ablation-execution} restricts execution pathways; Table~\ref{tab:ablation-priority} first replays a broader common candidate set and then varies the selection criterion.

\begin{table*}[t]
\centering
\caption{Execution-information ablation. Each entry reports target success / mean candidate evaluations.}
\label{tab:ablation-execution}
\small
\setlength{\tabcolsep}{3pt}
\resizebox{\textwidth}{!}{%
\begin{tabular}{lrrrrrrrr}
\toprule
Variant & Wikipedia & Reddit & MOOC & LastFM & CTDG avg. & ICEWS14 & ICEWS18 & TKG avg.\\
\midrule
Original execution only & 4.3\% / 34.3 & 8.7\% / 35.0 & 24.7\% / 34.7 & 16.7\% / 35.1 & 13.6\% / 34.8 & 4.7\% / 1.4 & 1.7\% / 0.9 & 3.2\% / 1.2 \\
Foil execution only & 3.3\% / 17.2 & 6.3\% / 14.8 & 21.7\% / 20.5 & 8.7\% / 7.9 & 10.0\% / 15.1 & 56.3\% / 17.0 & 82.0\% / 19.0 & 69.2\% / 18.0 \\
Complete execution contrast & 26.3\% / 42.2 & 32.3\% / 42.2 & 72.0\% / 45.6 & 48.3\% / 46.5 & 44.8\% / 44.1 & 64.0\% / 21.5 & 85.7\% / 21.7 & 74.8\% / 21.6 \\
\bottomrule
\end{tabular}}
\end{table*}

\begin{table*}[t]
\centering
\caption{Replay-based priority ablation. Each entry reports target success / mean evaluations including screening.}
\label{tab:ablation-priority}
\small
\setlength{\tabcolsep}{3pt}
\resizebox{\textwidth}{!}{%
\begin{tabular}{lrrrrrrrr}
\toprule
Variant & Wikipedia & Reddit & MOOC & LastFM & CTDG avg. & ICEWS14 & ICEWS18 & TKG avg.\\
\midrule
Original-score decrease & 12.0\% / 110.0 & 18.0\% / 102.9 & 42.3\% / 90.4 & 17.7\% / 97.3 & 22.5\% / 100.1 & 64.7\% / 53.4 & 84.3\% / 53.2 & 74.5\% / 53.3 \\
Foil-score increase & 29.7\% / 111.6 & 41.3\% / 106.1 & 78.7\% / 99.1 & 57.7\% / 102.8 & 51.8\% / 104.9 & 66.7\% / 53.6 & 86.0\% / 53.6 & 76.3\% / 53.6 \\
Contrastive-gap reduction & 27.0\% / 110.0 & 39.0\% / 104.5 & 71.7\% / 95.7 & 53.0\% / 101.0 & 47.7\% / 102.8 & 67.3\% / 53.6 & 86.3\% / 53.5 & 76.8\% / 53.5 \\
\bottomrule
\end{tabular}}
\end{table*}

In the first comparison, CTDG original-only generates DELETE/SHIFT from original pressure, whereas foil-only uses signed foil pressure and near-miss repair. Neither single-side variant uses inverse-transition repair; the complete variant includes this additional path. The increase from 13.6/10.0\% to 44.8\% therefore measures the \emph{combined effect of all execution-grounded generation pathways}, not the addition of a single information bit with every other candidate held fixed. TKG fixes the full candidate and composition sets, retaining original-trace DELETE/SHIFT for original-only and prefix-derived candidates for foil-only. The full method's original-terminal REWIRE is not wholly included in these single-side definitions. We interpret the two adapters' ablations according to these scopes.

In TKG, foil-side candidates alone attain 56.3/82.0\% success, compared with 64.0/85.7\% for complete execution. Unlike CTDG, the foil-side path supplies most successes. Together, these results support complementarity between completed original executions and foil-directed repair, not the logical necessity of both information sources in every case.

The second comparison computes candidate score changes through actual replay. CTDG uses the union of candidates from the execution variants; TKG uses a common trace-and-local candidate space. Each objective selects at most 16 atomic candidates and composes them with beam 8 for CTDG and beam 4 for TKG. CTDG complete-execution success of 44.8\% and replay-gap success of 47.7\% are not repeated measurements of the same priority. The former examines trace-prioritized candidates; the latter first replays a broader space and selects by observed effects. Mean CTDG evaluations including screening also differ: 44.1 versus 102.8.

Foil-score increase achieves 51.8\% in CTDG, exceeding gap reduction at 47.7\%; in TKG, gap reduction is highest at 67.3/86.3\%. Both foil-aware criteria outperform original-score decrease, but the best objective differs by backbone. We do not claim that the proposed \(\widehat R\) is more effective or accurate than all these replay-based objectives.

Candidate-cap and composition-beam sensitivity and operator-removal results are reported in Supplementary Section~\ref{supp:extended-ablation} and Tables~\ref{tab:ablation-width}--\ref{tab:ablation-operator}. Greater width improves success but raises composition cost. Increasing TKG beam from 8 to 16 adds 0.3 percentage points with approximately three times as many evaluations. Removed operators are not replaced by other candidates, so results quantify their unique successes within fixed candidate and composition sets. INSERT, REWIRE, and composition contribute in both formalisms, while the contributions of DELETE, SHIFT, and RELABEL vary across datasets.

\begin{figure}[t]
\centering
\includegraphics[width=0.96\columnwidth]{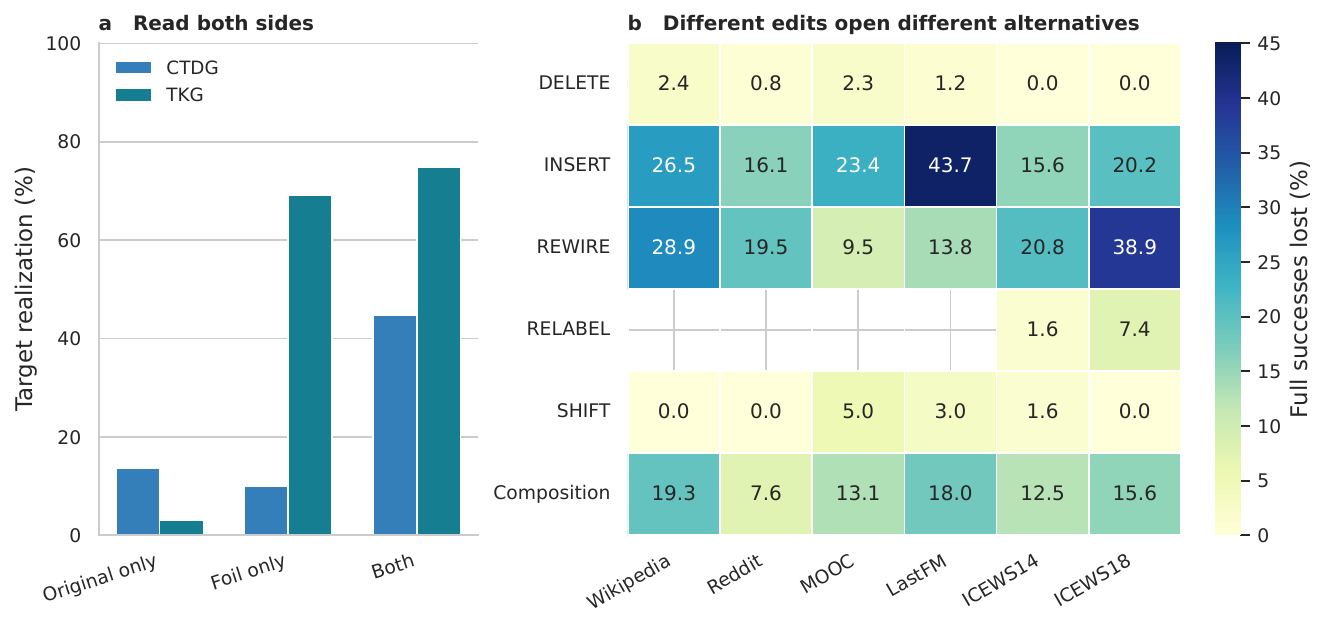}
\caption{Effects of combining execution-grounded generation pathways and contributions of operators/composition within fixed candidates. The specific restrictions imposed on single-side pathways are described in the text.}
\label{fig:mechanism-ablation}
\end{figure}

Overall, reusing execution involves both multiple proposal-generation pathways and actual replay. The complementarity in Fig.~\ref{fig:mechanism-ablation} supports their combination, rather than fully separating the independent causal effects of generation and search policy.

\subsection{Intervention-Space Diagnostics Across Temporal Graphs}\label{subsec:across-graphs}

We analyze whether small fact edits can reach designated foils in both temporal graph formalisms and which execution structures accompany difficulty. Every experiment fixes the original and rank-2/5/10 foils before intervention and records success only when the designated foil is returned as top-1 by the backbone's fixed inference. The CTDG results here are coordinate-based diagnostics of operator roles; the preceding efficiency experiment evaluates proposals generated from original--foil executions. The two experiments share queries, foils, and a maximum of two edits but answer different questions because their proposal-generation procedures differ.

\paragraph{Continuous-time dynamic graphs}\label{continuous-time-dynamic-graphs}

The LiFTER CTDG diagnostic evaluates 12,000 original--foil comparisons across four datasets. It constructs DELETE, INSERT, REWIRE, and SHIFT candidates by coordinate from the 24 most recent query-related events, up to 16 endpoints, and eight timestamp alternatives, then examines two-edit compositions from the top four candidates per operator. Within the fixed diagnostic space in Table~\ref{tab:ctdg-reachability}, mean reachability is 49.9\% for rank-2, 18.7\% for rank-5, and 9.2\% for rank-10 foils.

\begin{table}[t]
\centering
\caption{Bounded diagnostic reachability on continuous-time dynamic graphs.}
\label{tab:ctdg-reachability}
\small
\setlength{\tabcolsep}{4pt}
\begin{tabular}{lrrrr}
\toprule
Dataset & Rank-2 & Rank-5 & Rank-10 & Overall \\
\midrule
Wikipedia & 21.2\% & 6.3\% & 2.4\% & 10.0\% \\
Reddit & 20.2\% & 5.2\% & 1.1\% & 8.8\% \\
MOOC & 82.5\% & 28.4\% & 12.5\% & 41.1\% \\
LastFM & 75.8\% & 34.9\% & 20.6\% & 43.8\% \\
Average & 49.9\% & 18.7\% & 9.2\% & 25.9\% \\
\bottomrule
\end{tabular}
\end{table}

We examine declining CTDG reachability at the execution level. From rank 2 to rank 10, the number of grounded rules contributing to the foil decreases from 2.30 to 1.99 on Wikipedia, 2.42 to 2.02 on Reddit, 2.90 to 2.21 on MOOC, and 2.55 to 2.11 on LastFM. The foil's positive-rule-support deficit relative to the original increases on every dataset. In contrast, overlap between past-event rows used by the original and foil changes little. Lower reachability thus co-occurs with weaker executable support rather than markedly different event pools; the analysis does not identify a single cause of difficulty. Rank-2/5/10 provides difficulty strata with progressively weaker execution support.

We also separate operator and edit-budget roles in CTDG. A single DELETE reaches 19.3\%, 2.6\%, and 1.1\% of rank-2, rank-5, and rank-10 foils, respectively. Allowing all four operators raises single-edit coverage to 43.5\%, 12.6\%, and 5.9\%. Mean coverage across operators increases from 20.6\% with one edit to 25.9\% with at most two; 5.3\% of all cases require two-edit composition within this space. These results motivate a solver that handles structural and temporal changes alongside deletion and explores low-cost composition for alternatives beyond atomic reach.

The 10.0/8.8/41.1/43.8\% values in this table measure \textbf{bounded reachability} under coordinate-based diagnostic candidates, operators, and budgets. The earlier efficiency table's 30.8/37.1/77.2/57.0\% values measure \textbf{target success of the proposed method}, which constructs additional repair candidates from completed original and incomplete foil executions and retains up to 32. Both candidate construction and limiting procedures differ; both experiments use seed 7. Their numerical difference is therefore not attributed solely to generation. The diagnostic identifies useful types of changes, while the efficiency experiment evaluates how efficiently execution contrast finds them.

\paragraph{Temporal knowledge graphs}\label{temporal-knowledge-graphs}

The TLogic TKG experiment evaluates 600 original--foil comparisons on ICEWS14 and ICEWS18. Corresponding to CTDG interaction edits, we construct DELETE, INSERT, REWIRE, RELABEL, and SHIFT operations on timestamped facts and reapply TLogic rules to the edited knowledge graph.

\begin{table}[t]
\centering
\caption{Specified-foil success on temporal knowledge graphs.}
\label{tab:tkg-reachability}
\small
\setlength{\tabcolsep}{2.5pt}
\begin{tabular}{lrrrrrr}
\toprule
Dataset & Rank-2 & Rank-5 & Rank-10 & Overall & \makecell{One\\edit} & \makecell{Two\\edits} \\
\midrule
ICEWS14 & 73.0\% & 62.0\% & 57.0\% & 64.0\% & 168 & 24 \\
ICEWS18 & 90.0\% & 83.0\% & 84.0\% & 85.7\% & 217 & 40 \\
Combined & 81.5\% & 72.5\% & 70.5\% & 74.8\% & 385 & 64 \\
\bottomrule
\end{tabular}
\end{table}

In Table~\ref{tab:tkg-reachability}, TLogic makes the designated object top-1 in 449 of 600 comparisons (74.8\%). Of these, 385 require one fact edit and 64 require two-edit composition. Thus, beyond disrupting the original proof, interventions completing foil-side groundings and compositions of interventions produce successful TKG solutions.

The same computational principle operates on continuous interaction sequences and relational temporal facts. Original executions and incomplete foil executions are mapped backward to observable fact edits in each formalism, and exact replay verifies the specified alternative. Together with the efficiency results, this demonstrates a connection between concrete edits and replay across distinct representations.

\subsection{Target Specification Diagnostic}\label{why-the-foil-must-be-specified}

This diagnostic measures the empirical relationship between prediction invalidation and Specified-Foil Counterfactual success, not the performance of a particular existing explainer. We construct a controlled invalidation procedure that uses only the original \(A\)'s execution, without foil information. CTDG deletes past events supporting \(A\); TKG constructs DELETE and SHIFT from original traces. When \(A\) loses top-1, we check whether the new prediction equals a rank-2, rank-5, or rank-10 foil fixed beforehand. This measures how often changing \(A\) without targeting \(B\) happens to reach the designated alternative.

\begin{table}[t]
\centering
\caption{Where untargeted prediction changes arrive.}
\label{tab:target-specification}
\small
\setlength{\tabcolsep}{2.5pt}
\begin{tabular}{lrrrr}
\toprule
Formalism & \makecell{Changed\\predictions} & \makecell{Rank-2\\arrival} & \makecell{Rank-5\\arrival} & \makecell{Rank-10\\arrival} \\
\midrule
CTDG & 4,008 & 44.3\% & 3.6\% & 1.4\% \\
TKG & 122 & 91.8\% & 0.0\% & 0.0\% \\
\bottomrule
\end{tabular}
\end{table}

The high rank-2 arrival rates in Table~\ref{tab:target-specification} reflect the local ranking structure: the next-ranked candidate can readily succeed the original. Such incidental succession drops sharply for rank-5 and rank-10 foils in CTDG and never occurs in TKG. Prediction invalidation produces an outcome other than \(A\) without specifying which replacement. A Specified-Foil Counterfactual fixes \(B\) in the question and seeks changes that weaken original support or complete foil-side execution. Foil-directed reasoning therefore supplies the core information specifying which possible future is being investigated, rather than merely adding an evaluation criterion to an existing objective.

\subsection{Admissibility Sensitivity}\label{admissibility-sensitivity}

To test dependence on unrestricted event construction, we progressively strengthen validity through four policies. Protocol-valid enforces pre-query timestamps, entities within the model domain, and nonconflicting edits. Section~\ref{intervention-space} distinguishes the query-time fact from earlier occurrences of the same relation. No-INSERT additionally prohibits event creation. History-supported permits new CTDG $(source,destination)$ pairs or TKG $(subject,relation,object)$ triples introduced by INSERT, REWIRE, or RELABEL only if they occurred before the query. The final policy combines history support with the INSERT prohibition. Policies share queries, foils, candidate order, and composition frontiers; stricter policies only remove fixed candidates without replenishment.

\begin{table}[t]
\caption{Specified-foil reachability across all six datasets under stronger admissibility constraints. Retention is the fraction of the 1,039 protocol-valid successes preserved by each policy.}
\label{tab:admissibility-summary}
\centering
\footnotesize
\setlength{\tabcolsep}{2.5pt}
\begin{tabular}{lrr}
\toprule
Admissibility policy & Target success & Retention \\
\midrule
Protocol-valid & 1,039/1,800 (57.7\%) & 100.0\% \\
No INSERT & 790/1,800 (43.9\%) & 76.0\% \\
History-supported & 338/1,800 (18.8\%) & 32.5\% \\
History-supported, no INSERT & 301/1,800 (16.7\%) & 29.0\% \\
\bottomrule
\end{tabular}
\end{table}

In Table~\ref{tab:admissibility-summary}, prohibiting all INSERT operations retains 790/1,039 protocol-valid successes (76.0\%). Dataset-level retention ranges from 56.3\% to 84.4\%, showing that reaching the designated foil does not depend solely on event creation. Requiring historical precedent for every newly formed structure reduces retention to 32.5\%; additionally prohibiting INSERT yields 29.0\%. Creating new events and allowing new structures are distinct forms of flexibility. The former is unnecessary for most recovered solutions, whereas conservatively restricting the latter substantially contracts the reachable solution set.

History support is a conservative proxy permitting previously observed structures, not a comprehensive plausibility criterion. The framework's validity predicate applies application-defined admissibility before proposal generation and exact replay and reports reachability under that definition. This experiment, however, measures retention of already-generated proposals, not solver performance after regeneration and renewed search under stronger constraints. Dataset-level target success and retention relative to protocol-valid success are reported in Supplementary Section~\ref{supp:admissibility} and Table~\ref{tab:admissibility-by-dataset}.

\subsection{Qualitative Analysis}\label{subsec:qualitative}

To connect quantitative success to concrete edits and execution changes, we recover two representative cases with different operators from each formalism. Benchmark cases use rank-2/5 candidates as foils; the case compared with a real-world timeline uses the recorded ICEWS18 test answer. All foils are fixed before observing intervention outcomes, and only cases where replay returns the foil as top-1 are presented.

In MOOC, LiFTER predicts that student 348 will next interact with content 5993. We fix rank-5 content 6659 as the foil and ask what must differ for 6659 to be selected instead. The answer is deletion of one \texttt{(student\ 348,\ content\ 5981)} interaction occurring 18 time units before the query. Removing it decreases 5993's logit from 4.46 to 3.12 and increases 6659's from 2.13 to 3.57, making 6659 top-1. A single historical-interaction change thus reaches the designated alternative without inserting that alternative directly into the past.

The ICEWS18 test fact for September 28, 2018 records Benjamin Netanyahu as Abdel Fattah Al-Sisi's \texttt{Consult} target. TLogic instead ranks Moon Jae-in first and the correct answer, Netanyahu, ninth. We fix the observed answer as the foil and ask which preceding condition must change for Netanyahu to be predicted. No event addition or deletion is required: shifting the \texttt{(Benjamin\ Netanyahu,\ Express\ intent\ to\ meet\ or\ negotiate,\ Abdel\ Fattah\ Al-Sisi)} fact from September 24 to September 27, the day before the query, suffices. Exact replay increases Netanyahu's score from 0.286 to 0.738 and rank from ninth to first, while Moon Jae-in drops to second.

Day-by-day replay keeps Netanyahu ninth on September 25 and 26 and makes him top-1 only on September 27. Alongside contemporary reporting that the two leaders met in New York on the night of September 26~\cite{reuters2018sisi}, this identifies when the learned temporal dependency enables an alternative prediction. The external report is not treated as causal evidence for the intervention; the result is a temporal condition verified within the frozen predictor. Additional CTDG/TKG cases appear in Supplementary Section~\ref{supp:cases}.

\section{Simulator-level Validation of Model-derived Counterfactual Interventions}
\label{sec:simulator-validation}

Specified-Foil Counterfactuals do more than induce a model to produce a desired answer. They construct intervention hypotheses from model execution that can be tested outside the predictor, where separately executed simulator dynamics can confirm or challenge them.

A Specified-Foil Counterfactual already returns an executable intervention that changes the past to obtain a designated alternative, rather than merely a descriptive explanation of a prediction. Our primary evaluation establishes model-level validity by replaying the frozen predictor on the edited history and checking that the specified foil becomes top-1. Here, we extend the verification boundary to assess whether the same edit constructed from predictor execution produces the same foil in dynamics independent of that predictor.

We use the Pulse physiology simulator, which dynamically models battlefield casualties~\cite{bray2019pulse}. We fix \texttt{Survival} as the foil for casualties whose untreated Pulse trajectories end in \texttt{Death} and for whom LiFTER also predicts \texttt{Death}. The method reads LiFTER's \texttt{Death}--\texttt{Survival} execution contrast backward to construct an admissible treatment and its timing. If exact replay in LiFTER predicts \texttt{Survival}, we initialize Pulse with the same patient and injuries and execute the same treatment at the same time. Pulse uses none of LiFTER's scores, traces, or parameters and recomputes the subsequent physiological trajectory and terminal outcome from the initial state.

\subsection{Experimental Design}

The initial population comprises 8,687 public casualty specifications. Newly executed simulations use only demographic and injury specifications, not the public file's existing visits, serialized actions, or outcomes. For each casualty, we generate one untreated branch and branches applying injury-compatible treatments at four distinct times. Admissible treatments are tourniquets for limb hemorrhage, wound packing for abdominal hemorrhage, airway repositioning for airway obstruction, and needle decompression for tension pneumothorax. Sex, age, physical characteristics, and original injuries remain fixed; only treatment type and time are editable.

We materialized data for 894 newly simulated casualties and 6,083 legacy casualties used to augment training. Splits are casualty-level, preventing branches of the same patient from appearing in both training and evaluation. Validation and test contain only newly generated Pulse trajectories. LiFTER observes initial patient conditions, injuries, and optional treatments with their times. Post-treatment vital signs are descendants of the intervention and are therefore excluded from predictor inputs and recomputed by Pulse.

All branches predict \texttt{Death} or \texttt{Survival} at a common horizon of 75 minutes. Factual death times are retained as metadata for intervention-validity checks. Treatment times use a predetermined one-minute temporal vocabulary; for example, \texttt{treatment:tourniquet:m15} denotes a tourniquet applied within the 15-minute bin. This representation allows LiFTER to learn how treatment--time combinations contribute to terminal outcomes.

For this application, execution contrast and injury-specific admissibility define a finite set of treatment--time predicates, all of which are evaluated by exact replay. Pulse therefore tests transfer of model-valid interventions to an external simulator, rather than re-evaluating the main benchmark's search efficiency. Only candidates preceding factual death are allowed. Among those making \texttt{Survival} top-1, the solver returns the single \texttt{INSERT} with the largest \texttt{Survival} $-$ \texttt{Death} margin. Direct outcome insertion and changes to original injuries or patient conditions are prohibited.

Seed-7 LiFTER was trained for 10 epochs, achieving held-out outcome AUC 0.993 and accuracy 0.943. Eight held-out test casualties had untreated Pulse trajectories ending in \texttt{Death}; LiFTER predicted \texttt{Death} for seven. The method constructed model-level \texttt{Death} $\rightarrow$ \texttt{Survival} interventions for six of these seven. All six were evaluated in independently initialized Pulse simulations, without selecting only favorable examples.

\begin{table*}[t]
\centering
\caption{Model-derived counterfactual interventions replayed in Pulse physiology simulator.}
\label{tab:pulse-downstream}
\small
\setlength{\tabcolsep}{6pt}
\resizebox{\textwidth}{!}{%
\begin{tabular}{rlclcc}
\toprule
Casualty & Injury & Untreated death & Model-derived intervention & Model exact replay & Simulator replay \\
\midrule
601 & Left-leg hemorrhage, severity 2 & 45.0 min & Tourniquet at 15.5 min & Survival & \textbf{Survival} \\
885 & Right-arm hemorrhage, severity 3 & 27.0 min & Tourniquet at 15.5 min & Survival & \textbf{Survival} \\
910 & Right-leg hemorrhage, severity 5 & 15.0 min & Tourniquet at 0.5 min & Survival & \textbf{Survival} \\
934 & Liver hemorrhage, severity 4 & 64.0 min & Wound packing at 6.5 min & Survival & Death \\
1415 & Right-arm hemorrhage, severity 3 & 27.0 min & Tourniquet at 15.5 min & Survival & \textbf{Survival} \\
1419 & Splenic hemorrhage, severity 5 & 59.0 min & Wound packing at 12.5 min & Survival & \textbf{Survival} \\
\bottomrule
\end{tabular}%
}
\end{table*}

\subsection{Simulator-confirmed Interventions}

As summarized in Table~\ref{tab:pulse-downstream}, all replays completed without engine errors. LiFTER correctly predicted seven of eight untreated Pulse deaths and found \texttt{Death} $\rightarrow$ \texttt{Survival} interventions for six. Re-executing all six interventions in Pulse produced survival through 75 minutes in five independently computed trajectories. Thus, five of six model-level successes were confirmed by the simulator.

The ratio 5/6 is a proof-of-concept of executing and verifying model-derived interventions in independent simulator dynamics, not a clinical estimate of population-level treatment effectiveness.

Casualty 601 died at 45 minutes from left-leg hemorrhage without treatment. LiFTER proposed a tourniquet at 15.5 minutes, producing a \texttt{Survival} $-$ \texttt{Death} replay margin of $+15.885$. Executing the same treatment in Pulse yielded survival through 75 minutes. The proposed time was not among the four previously simulated treatment times, so this was not a retrieved successful branch.

Casualty 885 died at 27 minutes from right-arm hemorrhage. The proposed tourniquet at 15.5 minutes changed LiFTER's margin to $+15.873$ and produced 75-minute survival in Pulse. Casualty 1415 also died at 27 minutes from right-arm hemorrhage; the same 15.5-minute tourniquet led both LiFTER and Pulse to return \texttt{Survival}.

Casualty 910 had severe right-leg hemorrhage with severity 5 and died after 15 minutes without treatment. The earliest tourniquet sampled during data generation was scheduled at 15.67 minutes, after death, and none of the four existing treatment branches succeeded. From learned execution, the method constructed a previously unobserved tourniquet intervention at 0.5 minutes. LiFTER's margin became $+10.554$, and the first Pulse execution of this new branch produced survival through 75 minutes. This case most directly illustrates construction of a new intervention confirmed by independent dynamics, rather than retrieval of a stored simulator outcome.

Casualty 1419 had severity-5 splenic hemorrhage and died at 59 minutes in the untreated trajectory. Admissibility restricted abdominal-hemorrhage treatment to wound packing, and execution contrast selected an intervention at 12.5 minutes. LiFTER's margin became $+5.342$; halving hemorrhage severity in Pulse yielded survival through 75 minutes. Existing surviving branches applied treatment at 16.66, 35.41, and 54.16 minutes, so this intervention was also confirmed at a new time.

\subsection{Simulator Disagreement as Model Audit}

Casualty 934 had severity-4 liver hemorrhage and died at 64 minutes without treatment. LiFTER predicted \texttt{Survival} after wound packing at 6.5 minutes, with an exact-replay margin of $+5.411$. After executing the same action, however, Pulse produced brain oxygen deficit at 28.38 minutes, cardiovascular collapse at 28.74 minutes, and irreversible state at 28.86 minutes.

Existing comparison branches with wound packing at 15.68, 34.43, and 53.18 minutes are recorded as surviving. These results alone do not establish physiological nonmonotonicity in which earlier treatment is more harmful, or isolate predictor error. Distinguishing the causes requires comparing initial states, actual action parameters, and replay-adapter mappings. The current result identifies a concrete audit target: disagreement between the model-level verdict and simulator-level outcome for casualty 934's 6.5-minute intervention.

The model in this case study was trained on Pulse-derived trajectories. Separately initialized replay verifies execution without using predictor scores or parameters, but does not establish generalization to physiological mechanisms different from those in training or to real patients. Moreover, the specified-foil and invalidation criteria coincide for binary outcomes, and all finite treatment--time candidates are replayed; this study therefore does not test the main benchmark's multicandidate distinction or search efficiency. Confirmation of five of all six proposals provides a proof-of-concept of \textbf{executing synthesized interventions outside the predictor and checking their outcomes}.

\section{Discussion}
\label{sec:discussion}

\subsection{From Explanation to Intervention}

We reuse the execution that computed a prediction, rather than explanatory text itself. Knowing which facts contributed to a score through which bindings and temporal conditions enables construction of values that disrupt or satisfy those conditions. The trace serves as an intermediate representation between predictor and intervention solver: the predictor supplies executed evidence, the solver synthesizes candidates from that representation, and replay determines their actual effects. This connection makes the faithfulness and specificity of explanations operational inputs to further computation.

\subsection{Why Executability Matters}

The four conditions in Section~\ref{subsec:execution-interface} support distinct computational links. Faithful execution ties explanations to actual score computation; addressable grounding connects abstract conditions to editable events and values. Foil-queryable execution exposes established evidence and recoverable unmet conditions for an alternative not yet selected. Intervention-closed replay checks edits with the same predictor. Broader condition traceability enables more varied proposals, but recovering partial groundings neither enumerates the complete edit space nor guarantees success.

\subsection{Neuro-Symbolic Models as Operational XAI}

Neuro-symbolic models merit serious consideration because their reasoning can be reused computationally as well as inspected by people. Our two implementations illustrate this role. Evidence returned by rules and groundings identifies intervention targets and values, and edited executions yield verifiable outcomes. This perspective extends XAI from interpreting existing answers to constructing executable hypotheses for specified alternatives. Other architectures may also expose this structure; our results do not claim that such computation is exclusive to a particular model family.

\subsection{Scope and Open Questions}

Exact replay establishes model-level validity within a frozen predictor. Real-world causal effects and event plausibility require domain knowledge and additional assumptions~\cite{pearl2009causality}. The unit-cost budget of 2, finite proposals, and bounded composition also preclude guarantees of global minimality or unreachability. Quantitative evaluation covers LiFTER and TLogic; applying the approach to other executable models requires condition-recovery and edit adapters.

Shared-candidate experiments compare execution-informed selection and screening costs. Section~\ref{sec:candidate-synthesis} separately compares coordinate-based and execution-grounded generation under fixed downstream search in CTDGs and TKGs. This measures the two procedures, including candidate construction and selection within the cap, rather than superiority over all trace-free generators. Budget-matched comparisons with stronger targeted search remain future work. Fixed-proposal admissibility filtering also differs from solver performance when candidates are regenerated under new constraints. Pulse provides a small case study transferring proposals from a model trained on the same simulator family to separately initialized runs; it does not establish clinical effectiveness or generalization to unknown real-world dynamics. Within these limits, the contribution is \textbf{a procedure for reusing executable explanations in concrete intervention computation, together with its observed computational value}.

\section{Conclusion}
\label{sec:conclusion}

We have reused executable traces as an intermediate representation for intervention synthesis as well as evidence for predictions. Trace-guided intervention search transforms completed original and foil executions and recovered unmet conditions into concrete temporal fact edits, then verifies the designated foil through bounded composition and exact replay. LiFTER's interaction executions and TLogic's relational groundings connect to the same diagnosis--synthesis--verification procedure while preserving their respective score aggregations.

Shared-candidate comparisons and end-to-end measurements show that execution-informed candidate prioritization reduces replay costs while retaining high success rates. Ablations reveal complementary roles for execution information, editable coordinates, and composition. Pulse cases demonstrate a pathway for confirming or challenging model-derived interventions through separate simulator runs. This evidence holds within the two backbones and the specified search and admissibility settings.

\textbf{Explanation need not be the endpoint of reasoning.} When the structure that computes a prediction is linked to concrete facts and can be executed again, it can support both understanding of an existing answer and computation of the conditions for constructing and testing another.

\section*{Declaration of competing interest}
The authors declare that they have no known competing financial interests or
personal relationships that could have appeared to influence the work reported
in this paper.

\section*{Acknowledgements}
The authors have no acknowledgements to declare.

\section*{Data availability}

All six datasets used in this study---Wikipedia, Reddit, MOOC, LastFM, ICEWS14, and ICEWS18---are publicly available. The repository provides scripts to download the source files and materialize the processed experimental inputs. Dataset URLs are reported in Supplementary Table~\ref{tab:data-availability}.

\section*{Code availability}

The implementation, experiment configurations, aggregation scripts, and plotting scripts used in this study are available at \url{https://github.com/SnowyPainter/cf-public}.

\bibliographystyle{IEEEtran}
\bibliography{references}

\clearpage
\twocolumn[{%
  \centering
  {\LARGE Supplemental Material\par}
  \vspace{0.5em}
  {\large From Explanations to Interventions: Execution-Guided Counterfactual Synthesis in Temporal Graphs\par}
  \vspace{1.5em}
}]
\markboth{Supplemental Material}%
{Yu and Ha: From Explanations to Interventions}
\appendices
\section{Implementation and Statistical Details}
\label{supp:implementation}

The public source of every dataset is listed in Table~\ref{tab:data-availability}.

\begin{table*}[!t]
\centering
\caption{Public dataset sources used in this study.}
\label{tab:data-availability}
\small
\setlength{\tabcolsep}{4pt}
\renewcommand{\arraystretch}{1.15}
\begin{tabular}{p{0.12\textwidth}p{0.22\textwidth}p{0.58\textwidth}}
\toprule
Domain & Dataset & Source \\
\midrule
CTDG & Wikipedia & \url{https://snap.stanford.edu/jodie/wikipedia.csv} \\
CTDG & Reddit & \url{https://snap.stanford.edu/jodie/reddit.csv} \\
CTDG & MOOC & \url{https://snap.stanford.edu/jodie/mooc.csv} \\
CTDG & LastFM & \url{https://snap.stanford.edu/jodie/lastfm.csv} \\
TKG & ICEWS14 & TLogic repository: \url{https://github.com/liu-yushan/TLogic} \\
TKG & ICEWS18 & TLogic repository: \url{https://github.com/liu-yushan/TLogic} \\
\bottomrule
\end{tabular}
\end{table*}

LiFTER is trained before and independently of counterfactual search, with parameters frozen in all intervention experiments. We use AdamW for 10 epochs, batch size 512, learning rate $4\times10^{-3}$, weight decay $10^{-5}$, dropout 0.1, and maximum history length 128. Additional architecture settings are listed in Table~\ref{tab:backbone-settings}.

\begin{table*}[t]
\centering
\caption{LiFTER frozen-backbone configuration.}
\label{tab:backbone-settings}
\small
\setlength{\tabcolsep}{3.5pt}
\resizebox{\textwidth}{!}{%
\begin{tabular}{lr}
\toprule
Component & Value \\
\midrule
Training events / epochs & 59,500 / 10 \\
Batch size / learning rate & 512 / 0.004 \\
Weight decay / dropout & \(10^{-5}\) / 0.1 \\
History / fact-context length & 128 / 8 \\
Hidden / transition dimension & 64 / 32 \\
Maximum grounding facts / rule length & 10 / 2 \\
Maximum three-hop paths & 32 \\
\bottomrule
\end{tabular}%
}
\end{table*}

Main results use the frozen seed-7 checkpoint. Stability to training initialization is assessed using independently trained checkpoints with seeds 7, 17, and 37. All three runs share the temporal split, 1,000 query rows, rank-2/5/10 foil protocol, and search seed. We perform 20,000 paired cluster-bootstrap replicates, resampling queries while retaining their three foils and three training seeds together, and use exact McNemar tests on paired binary outcomes as a supplementary analysis. Table~\ref{tab:stability} reports the full results.

\begin{table*}[t]
\centering
\caption{Statistical stability over independently trained LiFTER checkpoints.}
\label{tab:stability}
\small
\setlength{\tabcolsep}{3.5pt}
\resizebox{\textwidth}{!}{%
\begin{tabular}{lrrrrr}
\toprule
Dataset & Trace-guided target success & Trace $-$ random, 95\% CI & Trace $-$ locality, 95\% CI & Trace $-$ greedy, 95\% CI & Evaluation reduction vs.\ greedy, 95\% CI \\
\midrule
Wikipedia & 30.38 $\pm$ 0.45\% & +12.40 [11.34, 13.47] & +13.47 [12.17, 14.79] & $-$2.36 [$-$2.83, $-$1.88] & 75.54 [75.22, 75.84]\% \\
Reddit & 36.56 $\pm$ 1.71\% & +12.82 [11.84, 13.81] & +27.69 [25.88, 29.51] & $-$2.37 [$-$2.83, $-$1.91] & 79.80 [79.55, 80.06]\% \\
MOOC & 77.16 $\pm$ 2.38\% & +25.23 [24.08, 26.38] & +32.98 [31.11, 34.92] & $-$4.84 [$-$5.46, $-$4.24] & 77.23 [77.08, 77.38]\% \\
LastFM & 56.26 $\pm$ 0.74\% & +18.99 [17.84, 20.13] & +20.43 [19.23, 21.62] & $-$9.72 [$-$10.47, $-$8.99] & 74.87 [74.69, 75.05]\% \\
\bottomrule
\end{tabular}%
}
\end{table*}

\section{Foil Construction Details}
\label{supp:foil-construction}

Before adopting rank-controlled foils, we compare the ground-truth destination, historical best (the currently highest-ranked destination previously chosen by the source), and historical recent (the most recently chosen destination). Table~\ref{tab:foil-construction} reports the fraction of queries for which each construction provides a foil distinct from the original, together with its pre-intervention median rank.

\begin{table*}[t]
\centering
\caption{Availability and pre-intervention rank of benchmark foil constructions. Each cell reports availability / median rank.}
\label{tab:foil-construction}
\small
\setlength{\tabcolsep}{3.5pt}
\resizebox{\textwidth}{!}{%
\begin{tabular}{lrrrr}
\toprule
Foil construction & Wikipedia & Reddit & MOOC & LastFM \\
\midrule
Rank-2 / 5 / 10 & 100\% / 2, 5, 10 & 100\% / 2, 5, 10 & 100\% / 2, 5, 10 & 100\% / 2, 5, 10 \\
Ground truth & 30\% / 12 & 39\% / 6 & 62\% / 3 & 79\% / 159 \\
Historical best & 57\% / 3 & 59\% / 2 & 100\% / 2 & 100\% / 2 \\
Historical recent & 57\% / 3 & 59\% / 3 & 100\% / 3 & 100\% / 19.5 \\
\bottomrule
\end{tabular}%
}
\end{table*}

The ground truth equals the original when the model predicts correctly and is therefore not a foil for those queries. Even among incorrect predictions, its median rank ranges from 3 to 159 across datasets. Historical constructions require sufficient source history and mix different difficulty levels. Rank-2/5/10 is defined for every query and fixes the comparison positions.

\section{Search-Width and Operator Ablations}
\label{supp:extended-ablation}

Table~\ref{tab:ablation-width} reports the full sensitivity grid for atomic candidate cap and composition beam. Larger candidate caps reach more foils on every dataset, while increasing the beam yields gains that saturate. Increasing the TKG beam from 8 to 16 improves success by only 0.3 percentage points on each dataset but approximately triples mean evaluations. In CTDG, the same expansion adds 3.8 points while increasing mean evaluations from 44.1 to 113.9.

\begin{table*}[t]
\centering
\caption{Sensitivity to atomic-candidate cap and composition beam.}
\label{tab:ablation-width}
\small
\setlength{\tabcolsep}{3.5pt}
\resizebox{\textwidth}{!}{%
\begin{tabular}{lrrrrrrrr}
\toprule
Search setting & Wikipedia & Reddit & MOOC & LastFM & CTDG avg. & ICEWS14 & ICEWS18 & TKG avg. \\
\midrule
Candidate cap 4 & 5.0\% & 14.0\% & 32.7\% & 8.0\% & 14.9\% & 38.7\% & 48.3\% & 43.5\% \\
Candidate cap 8 & 13.7\% & 23.7\% & 56.0\% & 32.3\% & 31.4\% & 49.0\% & 63.3\% & 56.2\% \\
Candidate cap 16 & 26.3\% & 32.3\% & 72.0\% & 48.3\% & 44.8\% & 64.0\% & 85.7\% & 74.8\% \\
Candidate cap 32 & 28.3\% & 39.0\% & 74.7\% & 56.3\% & 49.6\% & 73.0\% & 90.7\% & 81.8\% \\
Composition beam 2 & 22.0\% & 30.7\% & 58.7\% & 34.7\% & 36.5\% & 62.7\% & 80.3\% & 71.5\% \\
Composition beam 4 & 24.3\% & 31.7\% & 62.7\% & 38.7\% & 39.3\% & 64.0\% & 85.7\% & 74.8\% \\
Composition beam 8 & 26.3\% & 32.3\% & 72.0\% & 48.3\% & 44.8\% & 65.7\% & 87.0\% & 76.3\% \\
Composition beam 16 & 27.3\% & 34.0\% & 76.7\% & 56.3\% & 48.6\% & 66.0\% & 87.3\% & 76.7\% \\
\bottomrule
\end{tabular}%
}
\end{table*}

Table~\ref{tab:ablation-operator} reports, by dataset, the fraction of full-method successes lost when each operator or composition is removed. Candidates are fixed beforehand and removed slots are not replenished, so each value measures successes uniquely enabled by that component within the fixed set.

\begin{table*}[t]
\centering
\caption{Unique contribution of intervention operators and composition.}
\label{tab:ablation-operator}
\small
\setlength{\tabcolsep}{3.5pt}
\resizebox{\textwidth}{!}{%
\begin{tabular}{lrrrrrrrr}
\toprule
Removed component & Wikipedia & Reddit & MOOC & LastFM & CTDG overall & ICEWS14 & ICEWS18 & TKG overall \\
\midrule
DELETE & 2.4\% & 0.8\% & 2.3\% & 1.2\% & 1.7\% & 0.0\% & 0.0\% & 0.0\% \\
INSERT & 26.5\% & 16.1\% & 23.4\% & 43.7\% & 28.1\% & 15.6\% & 20.2\% & 18.3\% \\
REWIRE & 28.9\% & 19.5\% & 9.5\% & 13.8\% & 15.4\% & 20.8\% & 38.9\% & 31.2\% \\
RELABEL & -- & -- & -- & -- & -- & 1.6\% & 7.4\% & 4.9\% \\
SHIFT & 0.0\% & 0.0\% & 5.0\% & 3.0\% & 2.7\% & 1.6\% & 0.0\% & 0.7\% \\
Composition & 19.3\% & 7.6\% & 13.1\% & 18.0\% & 14.2\% & 12.5\% & 15.6\% & 14.3\% \\
\bottomrule
\end{tabular}%
}
\end{table*}

INSERT, REWIRE, and composition yield solutions that other components cannot replace in both formalisms. DELETE and SHIFT contributions vary by dataset, and RELABEL supplies additional unique solutions in relation-explicit TKGs. The operators need not contribute equally: event existence, endpoints, relations, and time provide distinct coordinates through which different alternatives become reachable.

\section{End-to-End Efficiency Details}
\label{supp:e2e-efficiency}

End-to-end time includes common preparation (history lookup and original prediction), method-specific proposal construction, atomic and composition replay, and final predictor top-1 verification. CUDA is synchronized at each GPU timing boundary, and method order is rotated across query--foil comparisons. Datasets run sequentially to avoid GPU/CPU resource contention in measurements. The first five of each dataset's 100 queries warm up every method; they remain in raw results but are excluded from timing summaries. Each method in Table~\ref{tab:e2e-efficiency} therefore reports 285 paired comparisons from 95 queries and three foils.

Black-box greedy evaluates every candidate in the shared admissible atomic universe, selects by foil margin, and reuses screening results during subsequent composition search. The shared universe includes trace-proposed candidates, but their generation time is charged only to the proposed method. Locality and random use the same universe, candidate cap, edit budget, and exact-replay criterion. This conservatively grants the controls identical candidate access without charging them for trace construction.

\begin{table*}[t]
\caption{End-to-end efficiency across datasets. Success denotes target success, time is mean wall-clock seconds per comparison, and throughput is successful specified-foil comparisons per second. The first five queries per dataset are warm-up and excluded from timing summaries, leaving 285 comparisons per method.}
\label{tab:e2e-efficiency}
\centering
\scriptsize
\setlength{\tabcolsep}{3.5pt}
\begin{tabular}{llrrrr}
\toprule
Dataset & Method & Success (\%) & Mean time (s) & P95 time (s) & Successes/s \\
\midrule
\multirow{4}{*}{Wikipedia} & Proposed & 27.7 & 0.590 & 2.219 & 0.470 \\
 & Black-box greedy & 30.9 & 1.254 & 3.229 & 0.246 \\
 & Locality & 11.9 & 0.260 & 0.987 & 0.460 \\
 & Random & 17.2 & 0.454 & 1.762 & 0.379 \\
\midrule
\multirow{4}{*}{Reddit} & Proposed & 42.8 & 0.840 & 2.542 & 0.510 \\
 & Black-box greedy & 47.0 & 2.026 & 4.288 & 0.232 \\
 & Locality & 14.7 & 0.342 & 1.069 & 0.431 \\
 & Random & 28.1 & 0.673 & 2.023 & 0.417 \\
\midrule
\multirow{4}{*}{MOOC} & Proposed & 73.7 & 0.220 & 0.443 & 3.355 \\
 & Black-box greedy & 77.5 & 0.702 & 1.379 & 1.104 \\
 & Locality & 33.7 & 0.130 & 0.213 & 2.590 \\
 & Random & 47.7 & 0.219 & 0.408 & 2.177 \\
\midrule
\multirow{4}{*}{LastFM} & Proposed & 60.7 & 0.549 & 1.169 & 1.105 \\
 & Black-box greedy & 70.9 & 0.999 & 1.864 & 0.709 \\
 & Locality & 37.5 & 0.236 & 0.542 & 1.593 \\
 & Random & 41.1 & 0.402 & 0.855 & 1.021 \\
\midrule
\multirow{4}{*}{ICEWS14} & Proposed & 64.2 & 8.317 & 13.122 & 0.077 \\
 & Black-box greedy & 67.7 & 11.171 & 15.923 & 0.061 \\
 & Locality & 56.1 & 6.396 & 8.863 & 0.088 \\
 & Random & 55.4 & 6.456 & 8.844 & 0.086 \\
\midrule
\multirow{4}{*}{ICEWS18} & Proposed & 84.9 & 73.304 & 121.072 & 0.012 \\
 & Black-box greedy & 85.6 & 101.282 & 169.224 & 0.008 \\
 & Locality & 62.1 & 57.571 & 98.332 & 0.011 \\
 & Random & 68.8 & 58.779 & 99.218 & 0.012 \\
\bottomrule
\end{tabular}
\end{table*}

The proposed method has lower mean and p95 latency than black-box greedy on every dataset. Throughput, defined as successful comparisons divided by total runtime, is also higher on all six datasets. Locality and random have lower absolute runtime on some datasets but reduced target success. Thus, fewer predictor evaluations translate into wall-clock savings, with trace-construction overhead below the cost of black-box atomic screening.

\section{Admissibility Sensitivity Details}
\label{supp:admissibility}

Admissibility sensitivity uses 100 queries with rank-2/5/10 foils for each of four CTDG and two TKG datasets. Strict policies mask the atomic candidates and composition frontier generated under the protocol-valid policy, without regeneration or reordering. Each strict reachable set is therefore a subset of the protocol-valid set. Bootstrap confidence intervals resample queries while keeping their three foils in one cluster.

History support checks whether a newly formed $(source,destination)$ pair in CTDG, or $(subject,relation,object)$ triple in TKG, occurred before the query. DELETE and SHIFT introduce no new structure. CTDG REWIRE and SHIFT preserve the edited event's feature vector, while INSERT uses a feature prototype from a grounded historical event. This tests structural precedent; clinical, social, or causal plausibility can be added through application-specific validity predicates.

\begin{table*}[t]
\caption{Admissibility sensitivity by dataset. Each cell reports target success and, in parentheses, retention of protocol-valid successes. All policies mask the same fixed candidates without replacement.}
\label{tab:admissibility-by-dataset}
\centering
\scriptsize
\setlength{\tabcolsep}{3.6pt}
\begin{tabular}{lrrrrrr}
\toprule
Policy & Wikipedia & Reddit & MOOC & LastFM & ICEWS14 & ICEWS18 \\
\midrule
Protocol-valid & 27.7 (100.0) & 39.3 (100.0) & 74.0 (100.0) & 55.7 (100.0) & 64.0 (100.0) & 85.7 (100.0) \\
No INSERT & 20.3 (73.5) & 33.0 (83.9) & 56.7 (76.6) & 31.3 (56.3) & 54.0 (84.4) & 68.0 (79.4) \\
History-supported & 5.3 (19.3) & 7.3 (18.6) & 37.0 (50.0) & 14.7 (26.3) & 17.0 (26.6) & 31.3 (36.6) \\
History-supported, no INSERT & 4.3 (15.7) & 7.3 (18.6) & 33.0 (44.6) & 12.3 (22.2) & 16.3 (25.5) & 27.0 (31.5) \\
\bottomrule
\multicolumn{7}{l}{\footnotesize Values are percentages: target success (retention among protocol-valid successes).}
\end{tabular}
\end{table*}

No-INSERT retention is 71.9\% across CTDG and 81.5\% across TKG. History-supported retention is nearly identical across formalisms, at 32.7\% and 32.3\%, respectively; combining history support with the INSERT prohibition retains 29.0\% in both. The observed trade-off is therefore not specific to one temporal graph representation. Once history support excludes most novel structures, additionally prohibiting INSERT produces only a small further reduction.

\section{Additional Qualitative Cases}
\label{supp:cases}

\subsection{Two-edit composition in a CTDG}

In LastFM, LiFTER predicts track 1176 for listener 1574, and rank-5 track 478 is designated as the foil. No atomic edit reaches 478. Adding a recent $(1574,1221)$ interaction and changing an older interaction's destination from 1172 to 707 makes 478 top-1. The two changes jointly satisfy distinct missing conditions, reaching an alternative beyond the examined single-edit space.

\subsection{Temporal boundary in a TKG}

In the ICEWS18 case, shifting the date one day at a time leaves Benjamin Netanyahu ninth on September 25 and 26 and makes him top-1 only on September 27. The contemporary report cited in the main paper states that Netanyahu and Abdel Fattah Al-Sisi met in New York on the night of September 26. We do not use this external record as causal evidence. Instead, juxtaposing the observed event timeline with the model's temporal boundary helps interpret how SHIFT changes TLogic's time-consistent grounding.

\subsection{A non-intuitive cross-relation dependency}

TLogic predicts \texttt{Citizen (Nigeria)} as Abdulrahman Dambazau's \texttt{Make an appeal or request} target, with \texttt{Nigeria} ranked second. Adding \texttt{(Abdulrahman Dambazau, Sexually assault, Nigeria)} on the day before the query makes Nigeria top-1. The added fact supplies a missing premise of a learned cross-relation rule, completing a grounding with Nigeria as its conclusion. Even a non-intuitive result is exposed as a chain of added fact, executed rule, and changed prediction, providing a concrete target for model audit.

\end{document}